\documentclass[runningheads]{llncs}
\usepackage{graphicx}
\usepackage{subfig}
\usepackage{amsmath,amssymb} 
\usepackage{color}
\usepackage[width=122mm,left=12mm,paperwidth=146mm,height=193mm,top=12mm,paperheight=217mm]{geometry}

\usepackage{multirow}
\usepackage{array}
\newcolumntype{L}[1]{>{\raggedright\let\newline\\\arraybackslash\hspace{0pt}}m{#1}}
\newcolumntype{C}[1]{>{\centering\let\newline\\\arraybackslash\hspace{0pt}}m{#1}}
\newcolumntype{R}[1]{>{\raggedleft\let\newline\\\arraybackslash\hspace{0pt}}m{#1}}

\begin{document}
\pagestyle{headings}
\mainmatter

\title{Multi-Scale Spatially-Asymmetric Recalibration for Image Classification} 

\titlerunning{Multi-Scale Spatially-Asymmetric Recalibration}

\authorrunning{Y. Wang, L. Xie, S. Qiao, Y. Zhang, W. Zhang and A. L. Yuille}

\author{Yan Wang\textsuperscript{1}, Lingxi Xie\textsuperscript{2}, Siyuan Qiao\textsuperscript{2},\\
Ya Zhang\textsuperscript{1}, Wenjun Zhang\textsuperscript{1}, Alan L. Yuille\textsuperscript{2}}

\institute{
\textsuperscript{1}Shanghai Jiao Tong University\quad
\textsuperscript{2}The Johns Hopkins University\\
}

\maketitle

\begin{abstract}
Convolution is {\em spatially-symmetric}, {\em i.e.}, the visual features are independent of its position in the image, which limits its ability to utilize contextual cues for visual recognition. This paper addresses this issue by introducing a {\em recalibration} process, which refers to the surrounding region of each neuron, computes an importance value and multiplies it to the original neural response. Our approach is named {\bf multi-scale spatially-asymmetric recalibration} (MS-SAR), which extracts visual cues from surrounding regions at {\em multiple scales}, and designs a weighting scheme which is {\em asymmetric in the spatial domain}. MS-SAR is implemented in an efficient way, so that only small fractions of extra parameters and computations are required. We apply MS-SAR to several popular building blocks, including the residual block and the densely-connected block, and demonstrate its superior performance in both CIFAR and ILSVRC2012 classification tasks.
\end{abstract}

\section{Introduction}
\label{Introduction}

In recent years, deep learning has been dominating in the field of computer vision. As one of the most important models in deep learning, the convolutional neural networks (CNNs) have been applied to various vision tasks, including image classification~\cite{krizhevsky2012imagenet}, object detection~\cite{girshick2015fast}, semantic segmentation~\cite{long2015fully}, boundary detection~\cite{xie2015holistically}, {\em etc}. The fundamental idea is to stack a number of linear operations ({\em e.g.}, convolution) and non-linear activations ({\em e.g.}, ReLU~\cite{nair2010rectified}), so that a deep network has the ability to fit very complicated distributions. There are two prerequisites in training a deep network, namely, the availability of large-scale image data, and the support of powerful computational resources.

Convolution is the most important operation in a deep network. A window is slid across the image lattice, and a number of small convolutional kernels are applied to capture local visual patterns. This operation suffers from a weakness of being {\em spatially-symmetric}, which assumes that visual features are independent of their spatial position. This limits the network's ability to learn from contextual cues ({\em e.g.}, an object is located upon another) which are often important in visual recognition. Conventional networks capture such spatial information by stacking a number of convolutions and gradually enlarging the receptive field, but we propose an alternative solution which equips {\em each} neuron with the ability to refer to its contexts at multiple scales efficiently.

\begin{figure}[t]
\begin{center}
    \includegraphics[width=11.6cm]{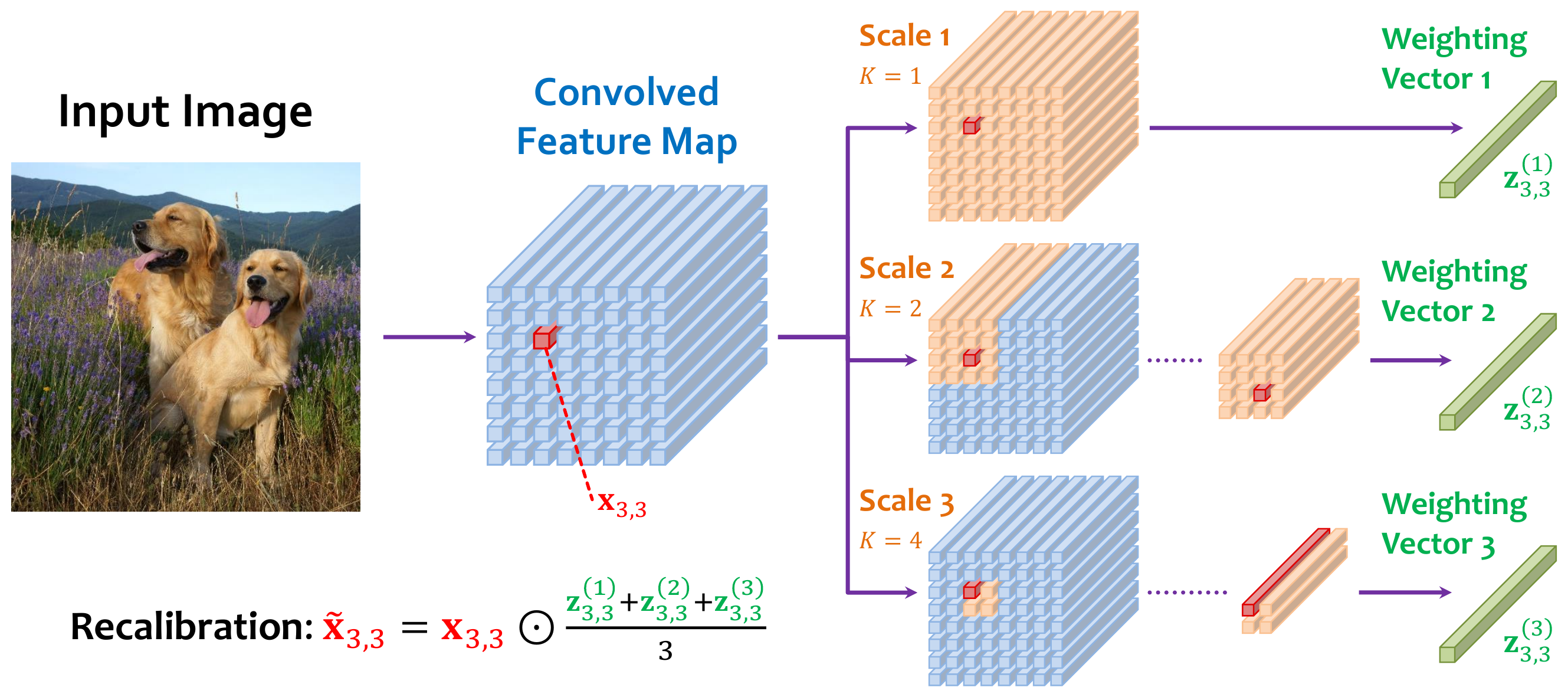}
\end{center}
\caption{
    Illustration of multi-scale spatially-asymmetric recalibration (MS-SAR). The feature vector for recalibration is marked in red, and the spatial coordinate sets at different scales are marked in yellow, and the weighting vectors are marked in green. For the first and second scales, for better visualization, we copy the neural responses used for recalibration. This figure is best viewed in color.
}
\label{Fig:Illustration}
\end{figure}

Our approach is named {\bf multi-scale spatially asymmetric recalibration} (MS-SAR). It quantifies the importance of each neuron by a score, and multiplies it to the original neural response. This process is named {\em recalibration}~\cite{hu2017squeeze}. Two features are proposed to enhance the effect of recalibration. First, the importance score of each neuron is computed from a local region (named a {\em coordinate set}) covering that neuron. This introduces the factor of spatial position into recalibration, leading to the desired {\em spatially-asymmetric} property. Second, we relate each neuron to multiple coordinate sets of different sizes, so that the importance of that neuron is evaluated by incorporating {\em multi-scale} information. The conceptual flowchart of our approach is illustrated in Figure~\ref{Fig:Illustration}.

In practice, the recalibration function (taking inputs from the coordinate sets and outputting the importance score) is the combination of two linear operations and two non-linear activations, and we allow the parameters to be learned from training data. To avoid heavy computational costs as well as a large amount of extra parameters to be introduced, we first perform a regional pooling over the coordinate set to reduce the spatial resolution, and use a smaller number of outputs in the first linear layer to reduce the channel resolution. Consequently, our approach only requires small fractions of extra parameters and computations beyond the baseline building blocks.

We integrate MS-SAR into two popular building blocks, namely, the residual block~\cite{he2016deep} and the densely-connected block~\cite{huang2017densely}, and empirically evaluate its performance in two image classification tasks. In the CIFAR datasets~\cite{krizhevsky2009learning}, our approach outperforms the baseline networks, the ResNets~\cite{he2016deep} and the DenseNets~\cite{huang2017densely}. In the ILSVRC2012 dataset~\cite{russakovsky2015imagenet}, we also compare with SENet~\cite{hu2017squeeze}, a special case of our approach with single-scale spatially-symmetric recalibration and demonstrate the superior performance of MS-SAR. In all cases, the extra computational overhead brought by MS-SAR does not exceed $1\%$.

The remainder of this paper is organized as follows. Section~\ref{RelatedWork} briefly reviews the previous literatures on image classification based on deep learning, and Section~\ref{Approach} illustrates the MS-SAR approach and describes how we apply it to different building blocks. After extensive experimental results are shown in Section~\ref{Experiments}, we conclude this work in Section~\ref{Conclusions}.

\section{Related Work}
\label{RelatedWork}

\subsection{Convolutional Neural Networks for Visual Recognition}
\label{RelatedWork:Recognition}

Deep convolutional neural networks (CNNs) have been widely applied to computer vision tasks. These models are based on the same motivation to learn and organize visual features in a hierarchical manner. In the early years, CNNs were verified successful in simple classification problems, in which the input image is small yet simple ({\em e.g.}, MNIST~\cite{lecun1998gradient} and CIFAR~\cite{krizhevsky2009learning}) and the network is shallow ({\em i.e.} with $3$--$5$ layers). With the emerge of large-scale image datasets~\cite{deng2009imagenet}\cite{lin2014microsoft} and powerful computational resources such as GPUs, it is possible to design and train deep networks for recognizing high-resolution natural images~\cite{krizhevsky2012imagenet}. Important technical advances involve using the piecewise-linear ReLU activation~\cite{nair2010rectified} to prevent under-fitting, and applying Dropout~\cite{srivastava2014dropout} to regularize the training process and avoid over-fitting.

Modern deep networks are built upon a handful of building blocks, including convolution, pooling, normalization, activation, element-wise operation (sum~\cite{he2016deep} or product~\cite{wang2017sort}), {\em etc}. Among them, convolution is considered the most important module to capture visual patterns by template matching (computing the inner-product between the input data and the learned templates), and most often, we refer to the depth of a network by the maximal number of convolutional layers along any path connecting the input to the output. It is believed that increasing the depth leads to better recognition performance~\cite{szegedy2015going}\cite{simonyan2015very}\cite{he2016deep}\cite{chen2017dual}\cite{huang2017densely}. In order to train these very deep networks efficiently, researchers proposed batch normalization~\cite{ioffe2015batch} to improve numerical stability, and highway connections~\cite{srivastava2015highway}\cite{he2016deep} to facilitate visual information to be propagated faster. The idea of automatically learning network architectures was also explored~\cite{xie2017genetic}\cite{zoph2017neural}.

Image classification lays the foundation of other vision tasks. The pre-trained networks can be used to extract high-quality visual features for image classification~\cite{donahue2014decaf}, instance retrieval~\cite{razavian2014cnn}, fine-grained object recognition~\cite{zhang2014part}\cite{xie2016interactive} or object detection~\cite{girshick2014rich}, surpassing the performance of conventional handcraft features. Another way of transferring knowledge learned in these networks is to fine-tune them to other tasks, including object detection~\cite{girshick2015fast}\cite{ren2015faster}, semantic segmentation~\cite{long2015fully}\cite{chen2016deeplab}, boundary detection~\cite{xie2015holistically}, pose estimation~\cite{toshev2014deeppose}\cite{newell2016stacked}, {\em etc}. A network with stronger classification results often works better in other tasks.

\subsection{Spatial Enhancement for Deep Networks}
\label{RelatedWork:SpatialEnhancement}

One of the most important factor of deep networks lies in the spatial domain. Although the convolution operation is naturally invariant to spatial translation, there still exist various approaches aimed at enhancing the ability of visual recognition by introducing different {\em priors} into deep networks.

In an image, the relationship between two features is often tighter when their spatial locations are closer to each other. An efficient way of modeling such distance-sensitive information is to perform spatial pooling~\cite{he2014spatial}, which explicitly splits the image lattice into several groups, and ignores the diversity of features in the same group. This idea is also widely used in object detection to summarize visual features given a set of regional proposals~\cite{girshick2015fast}\cite{ren2015faster}.

On the other hand, researchers also noticed that spatial importance (saliency) is not uniformly distributed in the spatial domain. Thus, various approaches were designed to discriminate the important (salient) features from others. Typical examples include using gradient back-propagation to find the neurons that contribute most to the classification result~\cite{zeiler2014visualizing}\cite{xie2016interactive}, introducing saliency~\cite{simonyan2013deep}\cite{noh2015learning} or attention~\cite{chen2016attention} into the network, and investigating local properties ({\em e.g.}, smoothness~\cite{xie2016geometric}). We note that a regular convolutional layer also captures local patterns in the spatial domain, but (i) it performs linear template matching and so cannot capture non-linear properties ({\em e.g.}, smoothness), meanwhile (ii) it often needs a larger number of parameters and heavier computational overheads.

In this work, we consider a {\em recalibration} approach~\cite{hu2017squeeze}, which aims at revising the response of each neuron by a spatial weight. Unlike~\cite{hu2017squeeze}, the proposed approach utilizes multi-scale visual information and allows different weights to be added at different spatial positions. This brings significant accuracy gains.

\section{Our Approach}
\label{Approach}

\subsection{Motivation: Why Spatial Asymmetry is Required?}
\label{Approach:Motivation}

Let $\mathbf{X}$ be the output of a convolutional layer. This is a 3D cube with $W\times H\times D$ entries, where $W$ and $H$ are the width and height, indicating the spatial resolution, and $D$ is the depth, indicating the number of convolutional kernels. According to the definition of convolution, each element in $\mathbf{X}$, denoted by $x_{w,h,d}$, represents the intensity of the $d$-th visual pattern at the coordinate $\left(w,h\right)$, which is obtained from the inner-product of the $d$-th convolutional kernel and the input region corresponding to the coordinate $\left(w,h\right)$.

Here we notice that convolution performs {\em spatially-symmetric} template matching, in which the intensity $x_{w,h,d}$ is independent of the spatial position $\left(w,h\right)$. We argue that this is not the optimal choice. In visual recognition, we often hope to learn contextual information ({\em e.g.}, feature $d_1$ often appears upon feature $d_2$), and so the {\em spatially-asymmetric} property is desired. To this end, we define $\mathcal{S}_{w,h}$ to be the {\em coordinate set} containing the neighboring coordinates of $\left(w,h\right)$ (detailed in the next subsection). We aim at computing a new response $\tilde{x}_{w,h,d}$ by taking into consideration all neural responses in $\mathcal{S}_{w,h}\times\left\{1,2,\ldots,D\right\}$, where $\times$ denotes the Cartesian product. Our approach is related but different from several existing approaches.

\begin{itemize}
\item First, we note that a standard convolution can learn contexts in a small local region, {\em e.g.}, $\mathcal{S}_{w,h}$ is a $3\times3$ square centered at $\left(w,h\right)$. Our approach can refer to multiple $\mathcal{S}_{w,h}$'s at different scales, capturing richer information and being more computationally efficient than convolution.
\item The second type works in the spatial domain, which uses the responses in the set $\mathcal{S}_{w,h}\times\left\{d\right\}$ to compute $\tilde{x}_{w,h,d}$. Examples include the Spatial Pyramid Pooling (SPP)~\cite{he2014spatial} layer which set regular pooling regions and ignored feature diversity within each region, and the Geometric Neural Phrase Pooling (GNPP)~\cite{xie2016geometric} layer which took advantage of the spatial relationship of neighboring neurons (it also assumed that spatially closer neurons have tighter connections) to capture feature co-occurrence. But, both of them are non-parameterized and work in each channel individually, which limited their ability to adjust feature weights.
\item Another related approach is called feature recalibration~\cite{hu2017squeeze}, which computed $\tilde{x}_{w,h,d}$ by referring to the visual cues in the entire image lattice, {\em i.e.}, the set $\left\{\left(w,h\right)\right\}_{w=1,h=1}^{W,H}\times\left\{1,2,\ldots,D\right\}$ was used. This is still a spatially-symmetric operation. As we shall see later, our approach is a generalized version and produces better visual recognition performance.
\end{itemize}

\subsection{Formulation: Spatially-Asymmetric Recalibration}
\label{Approach:Formulation}

Given the neural responses cube $\mathbf{X}$ and the coordinate set $\mathcal{S}_{w,h}$ at $\left(w,h\right)$, the goal is to compute a revised intensity $\tilde{x}_{w,h,d}$ with spatial information taken into consideration. We formulate it as a weighting scheme ${\tilde{x}_{w,h,d}}={x_{w,h,d}\times z_{w,h,d}}$, in which ${z_{w,h,d}}={f_d\!\left(\mathbf{X},\mathcal{S}_{w,h}\right)}$ and $f_d\!\left(\cdot\right)$ is named the {\em recalibration function}~\cite{hu2017squeeze}. This creates a weighting cube $\mathbf{Z}$ with the same size as $\mathbf{X}$ and propagate ${\tilde{\mathbf{X}}}={\mathbf{X}\odot\mathbf{Z}}$ to the next network layer. We denote the $D$-dimensional feature vector of $\mathbf{X}$ at $\left(w,h\right)$ by $\mathbf{x}_{w,h}={\left[x_{w,h,1};\ldots;x_{w,h,D}\right]^\top}$, and similarly for $\tilde{\mathbf{x}}_{w,h}$ and $\mathbf{z}_{w,h}$.

Let the set of all spatial positions be ${\mathcal{P}}={\left\{\left(w,h\right)\right\}_{w=1,h=1}^{W,H}}$. The coordinate set of each position is a subset of $\mathcal{P}$, {\em i.e.}, ${\mathcal{S}_{w,h}}\in{2^\mathcal{P}}$ where $2^\mathcal{P}$ is the power set of $\mathcal{P}$. Each coordinate set $\mathcal{S}_{w,h}$ defines a corresponding feature set ${\mathbf{X}_{\mathcal{S}_{w,h}}}={\left[\mathbf{x}_{w',h'}\right]_{\left(w',h'\right)\in\mathcal{S}_{w,h}}}$, and we abbreviate $\mathbf{X}_{\mathcal{S}_{w,h}}$ as $\mathfrak{X}_{w,h}$. Thus, ${z_{w,h,d}}={f_d\!\left(\mathbf{X},\mathcal{S}_{w,h}\right)}$ can be rewritten as ${z_{w,h,d}}={f_d\!\left(\mathfrak{X}_{w,h}\right)}$. This means that, for two spatial positions $\left(w_1,h_1\right)$ and $\left(w_2,h_2\right)$, $\mathbf{z}_{w_1,h_1}$ can be impacted by $\mathbf{x}_{w_2,h_2}$ if and only if ${\left(w_2,h_2\right)}\in{\mathcal{S}_{w_1,h_1}}$, and vice versa. It is common knowledge that if two positions $\left(w_1,h_1\right)$ and $\left(w_2,h_2\right)$ are close in the image lattice, {\em i.e.}, $\left\|\left(w_1,h_1\right)-\left(w_2,h_2\right)\right\|_1$ is small\footnote{Constraining $\left\|\left(w_1,h_1\right)-\left(w_2,h_2\right)\right\|_1$ results in a square region which is more friendly in implementation than constraining $\left\|\left(w_1,h_1\right)-\left(w_2,h_2\right)\right\|_2$.}, the relationship of their feature vectors is more likely to be tight. Therefore, we define each $\mathcal{S}_{w,h}$ to be a continuous region\footnote{By continuous we mean that $\mathcal{S}_{w,h}$ equals to the smallest convex hull that contains it, {\em i.e.}, there are no holes in this region.} that covers $\left(w,h\right)$ itself.

We provide two ways of defining $\mathcal{S}_{w,h}$, both of which are based on a scale parameter $K$. The first one is named the {\em sliding} strategy, in which ${\mathcal{S}_{w,h}}={\left\{\left(w',h'\right)\mid\left\|\left(w,h\right)-\left(w',h'\right)\right\|_1\leqslant T\right\}}$, where ${T}={\sqrt{WH}/K}$ is the threshold of distance. The second one is named the {\em regional} strategy, which partitions the image lattice into $K\times K$ equally-sized regions, and $\mathcal{S}_{w,h}$ is composed of all positions falling in the same region with it. The former is more flexible, {\em i.e.}, each position has a unique spatial region set, and so there are $W\times H$ different sets, while the latter reduces this number to $K^2$, which slightly reduces the computational costs (see Section~\ref{Approach:Costs}).

It remains to determine the form of the recalibration function $f_d\!\left(\mathfrak{X}_{w,h}\right)$. The major consideration is to reduce the number of parameters to alleviate the risk of over-fitting, and reduce the computational costs (FLOPs) to prevent the network from being much slower. We borrow the idea of adding both spatial and channel bottlenecks for this purpose~\cite{hu2017squeeze}. $\mathfrak{X}_{w,h}$ is first down-sampled into a single vector using average pooling, {\em i.e.}, ${\mathbf{y}_{w,h}}={\left|\mathcal{S}_{w,h}\right|^{-1}\sum_{\left(w,h\right)\in\mathcal{S}_{w,h}}\mathbf{x}_{w,h}}$, and passed through two fully-connected layers: ${z_{w,h,d}}={\sigma_2\!\left[\boldsymbol{\Omega}_{2,d}\cdot\sigma_1\!\left[\boldsymbol{\Omega}_1\cdot\mathbf{y}_{w,h}\right]\right]}$. Here, both $\boldsymbol{\Omega}_1$ and $\boldsymbol{\Omega}_{2,d}$ are learnable weight matrices, and $\sigma_1\!\left[\cdot\right]$ and $\sigma_2\!\left[\cdot\right]$ are activation functions which add non-linearity to the recalibration function. The dimension of $\boldsymbol{\Omega}_1$ is $D'\times D$ (${D'}<{D}$), and that of $\boldsymbol{\Omega}_{2,d}$ is $1\times D'$. This idea is similar to using channel bottleneck to reduce computations~\cite{he2016deep}. $\sigma_1\!\left[\cdot\right]$ is a composite function of batch normalization~\cite{ioffe2015batch} followed by ReLU activation~\cite{nair2010rectified}, and $\sigma_2\!\left[\cdot\right]$ replaces ReLU with sigmoid so as to output a floating point number in $\left(0,1\right)$.

We share $\boldsymbol{\Omega}_1$ over all $f_d\!\left(\mathfrak{X}_{w,h}\right)$'s, but use an individual $\boldsymbol{\Omega}_{2,d}$ for each output channel. Let ${\boldsymbol{\Omega}_2}={\left[\boldsymbol{\Omega}_{2,1}^\top;\ldots;\boldsymbol{\Omega}_{2,D}^\top\right]^\top}$, and thus the recalibration function is:
\begin{equation}
\label{Eqn:Recalibration}
{\mathbf{z}_{w,h}}={\mathbf{f}\!\left(\mathfrak{X}_{w,h}\right)}={\sigma_2\!\left[\boldsymbol{\Omega}_2\cdot\sigma_1\!\left[\boldsymbol{\Omega}_1\cdot\frac{1}{\left|\mathcal{S}_{w,h}\right|}\cdot\sum_{\left(w,h\right)\in\mathcal{S}_{w,h}}\mathbf{x}_{w,h}\right]\right]}.
\end{equation}

\subsection{Multi-Scale Spatially Asymmetric Recalibration}
\label{Approach:Implementation}

In Eqn~\eqref{Eqn:Recalibration}, the coordinate set $\mathcal{S}_{w,h}$ determines the region-of-interest (ROI) that can impact $\mathbf{z}_{w,h}$. There is the need of using different scales to evaluate the importance of each feature. We achieve this goal by defining multiple coordinate sets for each spatial position.

Let the total number of scales be $L$. For each ${l}={1,2,\ldots,L}$, we define the scale factor $K^{\left(l\right)}$, construct the coordinate set $\mathcal{S}_{w,h}^{\left(l\right)}$ and the feature set $\mathfrak{X}_{w,h}^{\left(l\right)}$, and compute $\mathbf{z}_{w,h}^{\left(l\right)}$ using Eqn~\eqref{Eqn:Recalibration}. The weights from different scales are averaged: ${\mathbf{z}_{w,h}}={\frac{1}{L}{\sum_{l=1}^L}\mathbf{z}_{w,h}^{\left(l\right)}}$. Using the matrix notation, we write {\bf multi-scale spatially-asymmetric recalibration} (MS-SAR) as:
\begin{equation}
\label{Eqn:MultiScaleRecalibration}
{\tilde{\mathbf{X}}_{w,h}}={\mathbf{X}\odot\mathbf{Z}}={\mathbf{X}\odot\frac{1}{L}{\sum_{l=1}^L}\mathbf{Z}^{\left(l\right)}}.
\end{equation}

The configuration of this an MS-SAR is denoted by ${\mathcal{L}}={\left\{K^{\left(l\right)}\right\}_{l=1}^L}$. When ${\mathcal{L}}={\left\{1\right\}}$, MS-SAR degenerates to the recalibration approach used in the Squeeze-and-Excitation Network (SENet)~\cite{hu2017squeeze}, which is single-scaled and spatially-symmetric, {\em i.e.}, each pair of spatial positions can impact each other, and $\mathbf{z}_{w,h}$ is the same at all positions. We will show in experiments that MS-SAR produces superior performance than this degenerated version.

\subsection{Applications to Existing Building Blocks}
\label{Approach:Applications}

\begin{figure}[t]
\begin{center}
    \includegraphics[width=11.0cm]{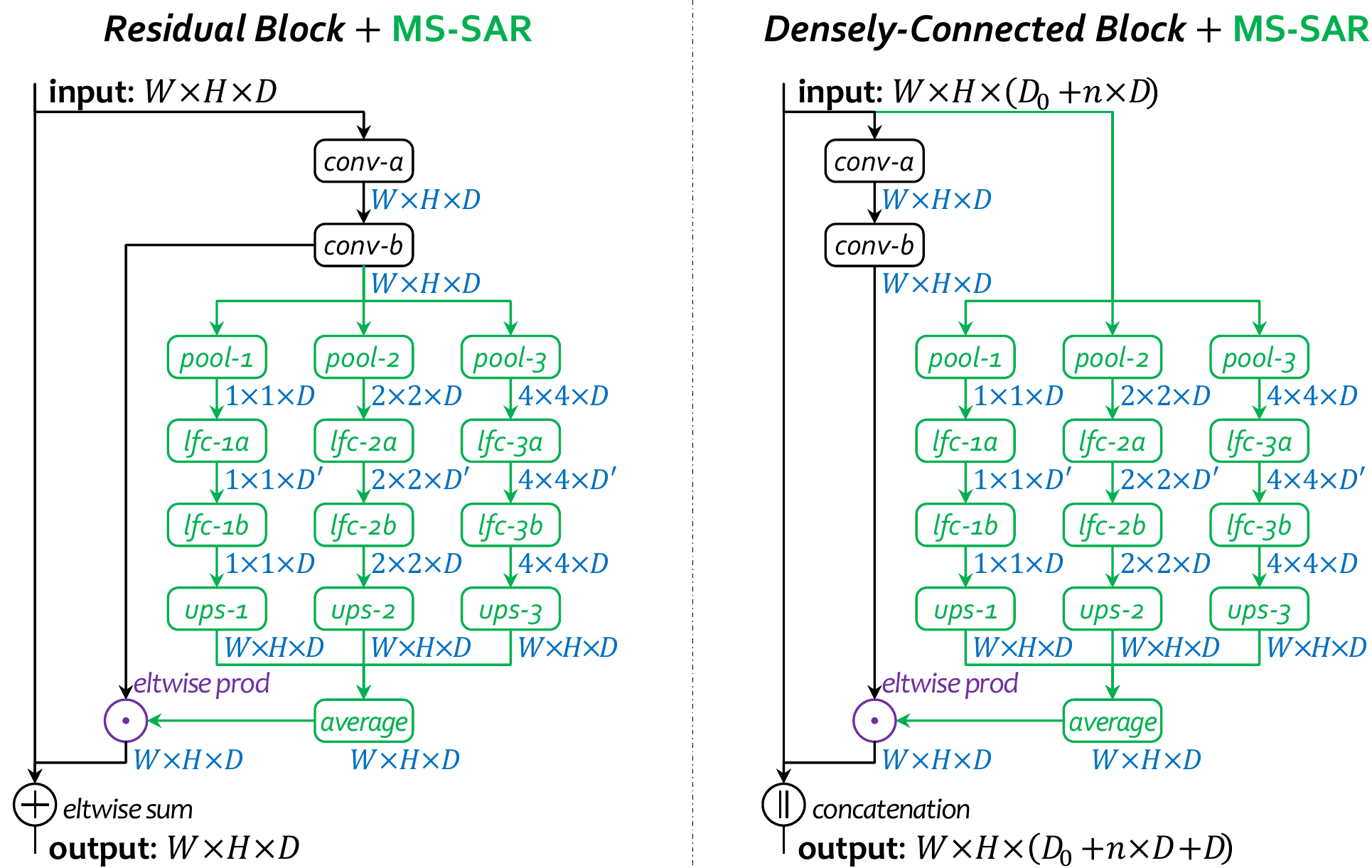}
\end{center}
\caption{
    Applying MS-SAR (green parts) to a residual block (left) or one single step in a densely-connected block (right). In both examples we set ${\mathcal{L}}={\left\{1,2,4\right\}}$. Here, {\em pool} indicates a $\frac{W}{K}\times\frac{H}{K}$ regional pooling, {\em lfc} is a local fully-connected layer ($1\times1$ convolution), and {\em ups} performs up-sampling by duplicating each element for $\frac{W}{K}\times\frac{H}{K}$ times. The feature map size is labeled for each cube. This figure is best viewed in color.
}
\label{Fig:Examples}
\end{figure}

MS-SAR can be applied to each convolutional layer individually. Here we consider two examples, which integrate MS-SAR into a residual block~\cite{he2016deep} and a densely-connected block~\cite{huang2017densely}, respectively. The modified blocks are shown in Figure~\ref{Fig:Examples}. In a residual block, we only recalibrate the second convolutional layer, while in a densely-connected block, this operation is performed before each convolved feature vector is concatenated to the main feature vector.

Another difference lies in the input of the recalibration function. In the residual block, we simply use the convolved response map for ``self recalibration'', but in the densely-connected block, especially in the late stages, we note that the main vector is of a much higher dimensionality and thus contains multi-stage visual information. Therefore, we compute the recalibration function using the main vector. We name this option as {\em multi-stage recalibration}. In comparison to {\em single-stage recalibration} (input the convolved vector to the recalibration function), it requires more parameters as well as computations, but also leads to better classification performance (see Section~\ref{Experiments:Ablation}).

\subsection{Computational Costs}
\label{Approach:Costs}

Let $\mathbf{X}$ be a $W\times H\times D$ cube, and the input of convolution also have $D$ channels, then the number of parameters of convolution is $9D^2$ (assuming the convolution kernel size is $3\times3$). Given that MS-SAR is configured by ${\mathcal{L}}={\left\{K^{\left(l\right)}\right\}_{l=1}^L}$, the learnable parameters come from two weight matrices $\boldsymbol{\Omega}_1$ ($D'\times D$) and $\boldsymbol{\Omega}_2$ ($D\times D'$), and so there are $2DD'$ extra parameters for each scale, and $2LDD'$ for all $L$ scales. We set ${D'}={D/L}$ so that using multiple scales does not increase the total number of parameters.

The extra computations (FLOPs) brought by MS-SAR is related to the strategy of defining the coordinate sets. We first consider the {\em sliding} strategy, in which each position $\left(w,h\right)$ has a different feature set $\mathfrak{X}_{w,h}$. The spatial average pooling over the feature sets of all positions takes around $WHD$ FLOPs\footnote{This is implemented by the idea of {\em partial sum}. For each channel, we compute ${T_{w,h}}={{\sum_{w'=1}^w}{\sum_{h'=1}^h}{\sum_{d=1}^D}x_{w',h',d}}$ for each position $\left(w,h\right)$ -- using a gradual accumulation process, this takes $WHD$ sum operations for all $D$ channels. Then we have ${{\sum_{w=w_1}^{w_2}}{\sum_{h=h_1}^{h_2}}{\sum_{d=1}^D}x_{w,h,d}}={T_{w_2,h_2}-T_{w_1-1,h_2}-T_{w_2,h_1-1}+T_{w_1-1,h_1-1}}$, which takes $O\!\left(WH\right)$ sum operations for all spatial position $\left(w,h\right)$'s.}. Then, each $D$-dimensional vector $\mathbf{y}_{w,h}$ is passed through two matrix-vector multiplications, and the total FLOPs is $2WHDD'$. For the {\em regional} strategy, the difference lies in that the number of unique feature sets is $K^{\left(l\right)2}$ at the $l$-th scale. By sharing computations, the total FLOPs of the fully-connected layers is decreased to $2K^{\left(l\right)2}DD'$. For all $L$ scales, the extra FLOPs is $2LWHDD'$ for the {\em sliding} strategy and $2DD'{\sum_{l=1}^L}K^{\left(l\right)2}$ for the {\em regional} strategy, respectively.

Note that in both ResNets and DenseNets, MS-SAR is applied to half of convolutional layers, and so the fractions of extra parameters and FLOPs are relatively small. We will report the detailed numbers in experiments.

\section{Experiments}
\label{Experiments}

\subsection{The CIFAR Datasets}
\label{Experiments:CIFAR}

\newcommand{\colwidthA}{1.2cm}
\newcommand{\colwidthB}{3.0cm}
\newcommand{\colwidthC}{4.56cm}
\begin{table*}[!btp]
\centering
\begin{tabular}{|C{\colwidthA}|c||C{\colwidthB}|C{\colwidthB}|C{\colwidthB}|}
\hline
Layer         & Size         & ResNet-20                 & ResNet-32                 & ResNet-56                 \\
\hline\hline
{\em conv-1}  & $32\times32$ & \multicolumn{3}{c|}{$3\times3$, $16$ (kernel size, channel \#)}                   \\
\hline
{\em conv-2}  & $32\times32$ & $\begin{bmatrix} 3\times3,   16 \\ 3\times3,   16 \end{bmatrix}\times 3$
                             & $\begin{bmatrix} 3\times3,   16 \\ 3\times3,   16 \end{bmatrix}\times 5$
                             & $\begin{bmatrix} 3\times3,   16 \\ 3\times3,   16 \end{bmatrix}\times 9$ \\
\hline
{\em conv-3}  & $16\times16$ & $\begin{bmatrix} 3\times3,   32 \\ 3\times3,   32 \end{bmatrix}\times 3$
                             & $\begin{bmatrix} 3\times3,   32 \\ 3\times3,   32 \end{bmatrix}\times 5$
                             & $\begin{bmatrix} 3\times3,   32 \\ 3\times3,   32 \end{bmatrix}\times 9$ \\
\hline
{\em conv-4}  & $ 8\times 8$ & $\begin{bmatrix} 3\times3,   64 \\ 3\times3,   64 \end{bmatrix}\times 3$
                             & $\begin{bmatrix} 3\times3,   64 \\ 3\times3,   64 \end{bmatrix}\times 5$
                             & $\begin{bmatrix} 3\times3,   64 \\ 3\times3,   64 \end{bmatrix}\times 9$ \\
\hline
{\em fc}      & $ 1\times 1$ & \multicolumn{3}{c|}{global average pooling, $1\times1$, $10$ or $100$, fully-connected} \\
\hline\hline
FLOPs         &              & $40.8\mathrm{M}$          & $69.1\mathrm{M}$          & $135.7\mathrm{M}$         \\
\hline
\end{tabular}
\begin{tabular}{|C{\colwidthA}|c||C{\colwidthC}|C{\colwidthC}|}
\hline
Layer         & Size         & DenseNet-100              & DenseNet-190               \\
\hline\hline
{\em conv-1}  & $32\times32$ & \multicolumn{2}{c|}{$3\times3$, $2\times\mathrm{growth}$ (kernel size, channel \#)}             \\
\hline
{\em dense-2} & $32\times32$ & $\begin{bmatrix} 1\times1,   48 \\ 3\times3,   12 \end{bmatrix}\times 16$; $\begin{matrix} \mathrm{base} & = & 24 \\ \mathrm{growth} & = & 12 \end{matrix}$
                             & $\begin{bmatrix} 1\times1,   160 \\ 3\times3,  40 \end{bmatrix}\times 31$; $\begin{matrix} \mathrm{base} & = & 80 \\ \mathrm{growth} & = & 40 \end{matrix}$ \\
\hline
{\em dense-3} & $16\times16$ & $\begin{bmatrix} 1\times1,   48 \\ 3\times3,   12 \end{bmatrix}\times 16$; $\begin{matrix} \mathrm{base} & = & 24 \\ \mathrm{growth} & = & 12 \end{matrix}$
                             & $\begin{bmatrix} 1\times1,   160 \\ 3\times3,  40 \end{bmatrix}\times 31$; $\begin{matrix} \mathrm{base} & = & 80 \\ \mathrm{growth} & = & 40 \end{matrix}$ \\
\hline
{\em dense-4} & $ 8\times 8$ & $\begin{bmatrix} 1\times1,   48 \\ 3\times3,   12 \end{bmatrix}\times 16$; $\begin{matrix} \mathrm{base} & = & 24 \\ \mathrm{growth} & = & 12 \end{matrix}$
                             & $\begin{bmatrix} 1\times1,   160 \\ 3\times3,  40 \end{bmatrix}\times 31$; $\begin{matrix} \mathrm{base} & = & 80 \\ \mathrm{growth} & = & 40 \end{matrix}$ \\
\hline
{\em fc}      & $ 1\times 1$ & \multicolumn{2}{c|}{global average pooling, $1\times1$, $10$ or $100$, fully-connected}         \\
\hline\hline
FLOPs         &              & $252.5\mathrm{M}$         & $7.9\mathrm{G}$           \\
\hline
\end{tabular}
\caption{
    Network architectures for the CIFAR datasets. Batch normalization~\cite{ioffe2015batch} and ReLU activation~\cite{nair2010rectified} are used as in the original papers~\cite{he2016deep}\cite{huang2017densely}.
}
\label{Tab:ArchitecturesCIFAR}
\end{table*}

We first evaluate MS-SAR on the CIFAR datasets~\cite{krizhevsky2009learning} which contain tiny RGB images with a fixed spatial resolution of $32\times32$. There are two subsets with $10$ and $100$ object classes, referred to as CIFAR10 and CIFAR100, respectively. Each set has $50\rm{,}000$ training samples and $10\rm{,}000$ testing samples, both of which are evenly distributed over all ($10$ or $100$) classes.

We choose different baseline network architectures, including the deep residual networks (ResNets)~\cite{he2016deep} with $20$, $32$ and $56$ layers and the densely-connected networks (DenseNets)~\cite{huang2017densely} with $100$ and $190$ layers. Detailed network architectures are shown in Table~\ref{Tab:ArchitecturesCIFAR}. MS-SAR is applied to {\em each} residual block and densely-connected block, as illustrated in Figure~\ref{Fig:Examples}. We choose the {\em regional} strategy to construct coordinate sets, use ${\mathcal{L}}={\left\{1,2,4\right\}}$ and set ${D'}={D/3}$. For other options, see ablation studies in the next subsection.

We follow the conventions to train these networks from scratch. We use the standard Stochastic Gradient Descent (SGD) with a weight decay of $0.0001$ and a Nesterov momentum of $0.9$. In the ResNets, we train the network for $160$ epochs with mini-batch size of $128$. The base learning rate is $0.1$, and is divided by $10$ after $80$ and $120$ epochs. In the DenseNets, we train the network for $300$ epochs with a mini-batch size of $64$. The base learning rate is $0.1$, and is divided by $10$ after $150$ and $225$ epochs. Adding MS-SAR does not require any of these settings to be modified. In the training process, the standard data-augmentation is used, {\em i.e.}, each image is padded with a $4$-pixel margin on each of the four sides. In the enlarged $40\times40$ image, a subregion with $32\times32$ pixels is randomly cropped and flipped with a probability of $0.5$. No augmentation is used at the testing stage.

\renewcommand{\colwidthA}{0.8cm}
\renewcommand{\colwidthB}{1.0cm}
\renewcommand{\colwidthC}{1.2cm}
\newcommand{\colwidthD}{1.2cm}
\begin{table*}[!btp]
\centering
\begin{tabular}{|l||R{\colwidthA}|R{\colwidthB}||l||R{\colwidthA}|R{\colwidthB}|R{\colwidthC}|R{\colwidthD}|}
\hline
Approach                                              &     C10 &    C100 & Network &              C10 &             C100 &             FLOPs &           Params \\
\hline\hline
Lee {\em et al.}, 2015~\cite{lee2015deeply}           & $ 7.97$ & $34.57$ & RN-20   &          $ 8.61$ &          $31.87$ & $ 40.8\mathrm{M}$ & $0.27\mathrm{M}$ \\
\hline
He {\em et al.}, 2016~\cite{he2016deep}               & $ 6.61$ & $27.22$ & RN-20*  & $\mathbf{ 7.61}$ & $\mathbf{31.09}$ & $ 40.9\mathrm{M}$ & $0.28\mathrm{M}$ \\
\hline
Huang {\em et al.}, 2016~\cite{huang2016deep}         & $ 5.23$ & $24.58$ & RN-32   &          $ 7.51$ &          $30.63$ & $ 69.1\mathrm{M}$ & $0.46\mathrm{M}$ \\
\hline
He {\em et al.}, 2016~\cite{he2016identity}           & $ 4.62$ & $22.71$ & RN-32*  & $\mathbf{ 6.68}$ & $\mathbf{29.41}$ & $ 69.3\mathrm{M}$ & $0.48\mathrm{M}$ \\
\hline
Zagoruyko {\em et al.}, 2016~\cite{zagoruyko2016wide} & $ 4.17$ & $20.50$ & RN-56   &           $6.97$ &          $29.07$ & $125.7\mathrm{M}$ & $0.85\mathrm{M}$ \\
\hline
Han {\em et al.}, 2017~\cite{han2017deep}             & $ 3.48$ & $17.01$ & RN-56*  & $\mathbf{ 6.04}$ & $\mathbf{27.71}$ & $126.0\mathrm{M}$ & $0.89\mathrm{M}$ \\
\hline
Huang {\em et al.}, 2017~\cite{huang2017snapshot}     & $ 3.40$ & $17.40$ & DN-100  &          $ 4.67$ &          $22.45$ & $252.5\mathrm{M}$ & $0.80\mathrm{M}$ \\
\hline
Zhang {\em et al.}, 2017~\cite{zhang2017interleaved}  & $ 3.25$ & $19.25$ & DN-100* & $\mathbf{ 4.16}$ & $\mathbf{21.13}$ & $253.3\mathrm{M}$ & $0.99\mathrm{M}$ \\
\hline
Gastaldi {\em et al.}, 2017~\cite{gastaldi2017shake}  & $ 2.86$ & $15.85$ & DN-190  &          $ 3.46$ &          $17.34$ & $ 7.95\mathrm{G}$ & $25.8\mathrm{M}$ \\
\hline
Zhang {\em et al.}, 2017~\cite{zhang2017mixup}        & $ 2.70$ & $16.80$ & DN-190* & $\mathbf{ 3.32}$ & $\mathbf{16.92}$ & $ 7.98\mathrm{G}$ & $32.7\mathrm{M}$ \\
\hline
\end{tabular}
\caption{
    Comparison of classification error rates ($\%$) on the CIFAR10 and CIFAR100 datasets. The left three columns list several recent work, and the right part compares our approach with the baselines. ``RN'' and ``DN'' denotes ``ResNet'' and ``DenseNet''. An asterisk sign (*) indicates that MS-SAR is added. For all ResNets, the error rates are averaged from $3$ individual runs. All FLOPs and numbers of parameters are computed on the experiments on CIFAR10. The difference in these numbers between the CIFAR10 and CIFAR100 experiments are ignorable.
}
\label{Tab:ResultsCIFAR}
\end{table*}

Classification results are summarized in Table~\ref{Tab:ResultsCIFAR}. One can observe that MS-SAR improves the baseline classification accuracy consistently and significantly. In particular, in terms of the relative drop in error rates, almost all these numbers are higher than $10\%$ on CIFAR10 (except for DenseNet-190), and higher than $4\%$ on CIFAR100 (except for ResNet-20 and DenseNet-190). The highest drop is over $10\%$ on CIFAR10 and over $5\%$ on CIFAR100. We note that these improvements are produced at the price of higher model complexities. The additional computational costs are very small for both the ResNets ({\em e.g}, $\sim0.3\%$ extra FLOPs) and DenseNets ({\em e.g}, $\sim0.3\%$ and $\sim0.4\%$ extra FLOPs for DenseNet-100 and DenseNet-190, respectively), and the fractions of extra parameters are moderate ($\sim5\%$ for the ResNets and $\sim25\%$ for the DenseNets, respectively).

We also compare our results with the state-of-the-arts (listed in the left part of Table~\ref{Tab:ResultsCIFAR}). Although some recent approaches reported much higher accuracies in the CIFAR datasets, we point out that they often used larger spatial resolutions~\cite{han2017deep}, complicated network modules~\cite{zhang2017interleaved} or complicated regularization methods~\cite{gastaldi2017shake}\cite{zhang2017mixup}, and thus the results are not directly comparable to ours. In addition, we believe that MS-SAR can be applied to these networks towards better classification performance.

\begin{figure}[t]
\begin{center}
    \includegraphics[width=4.0cm]{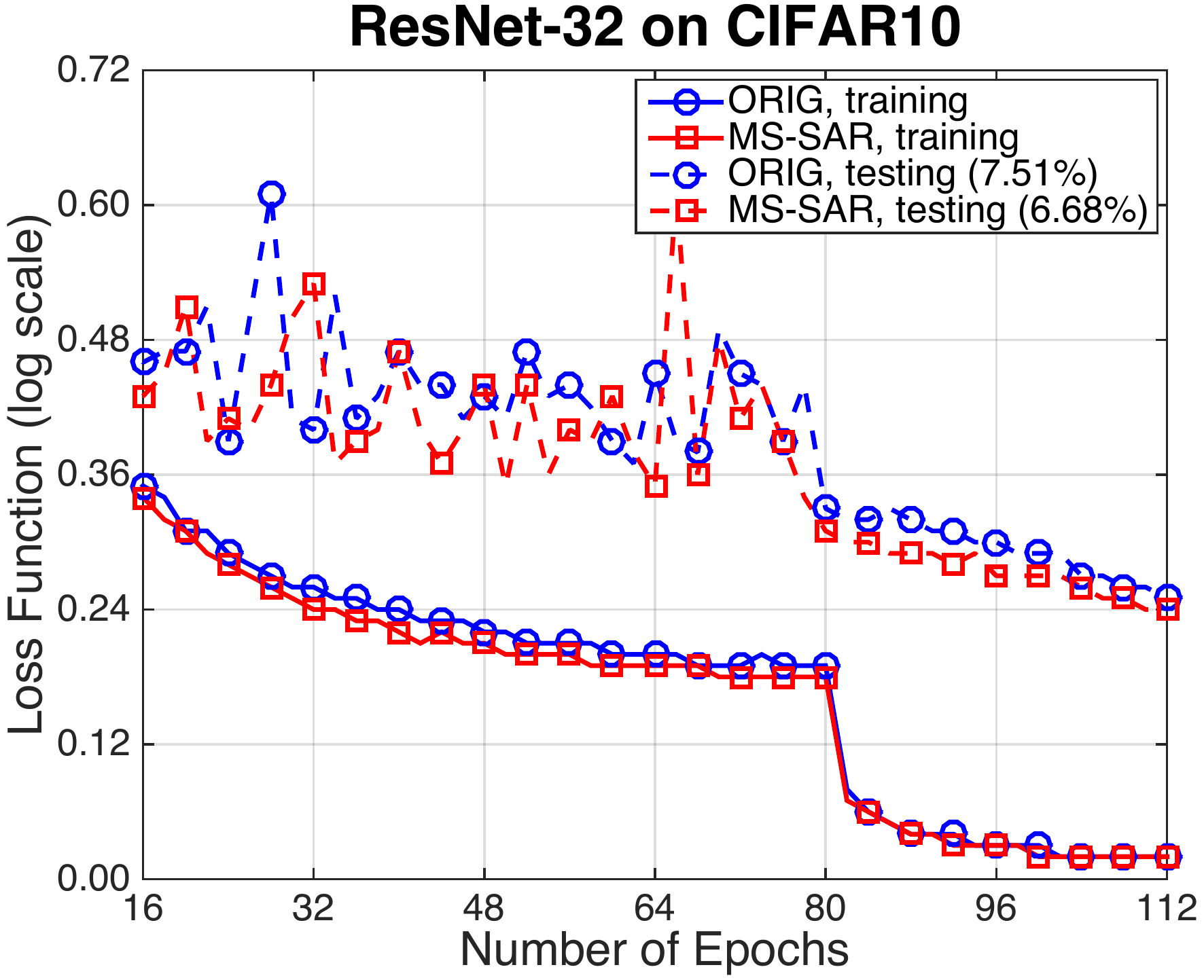}
    \includegraphics[width=4.0cm]{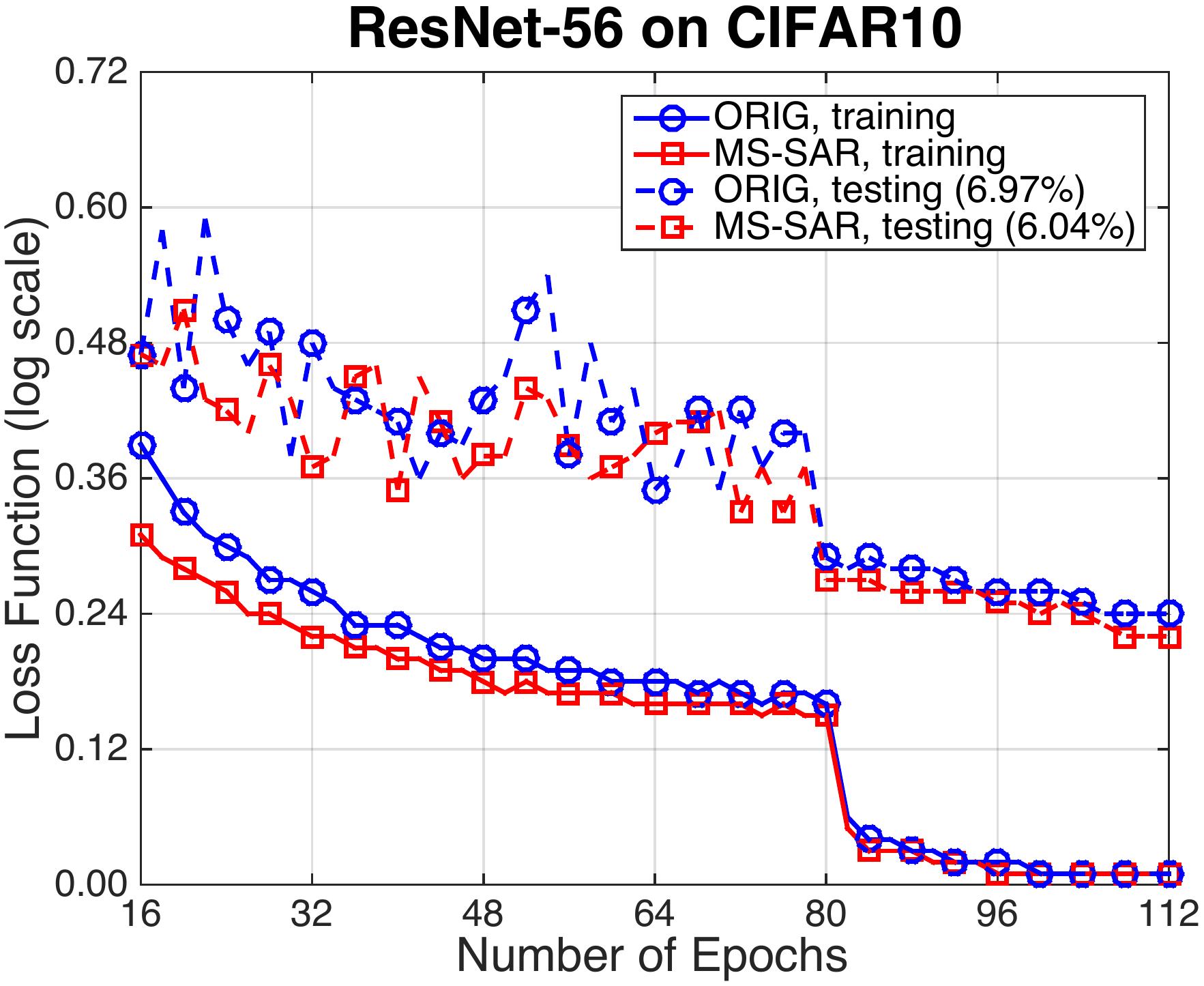}
    \includegraphics[width=4.0cm]{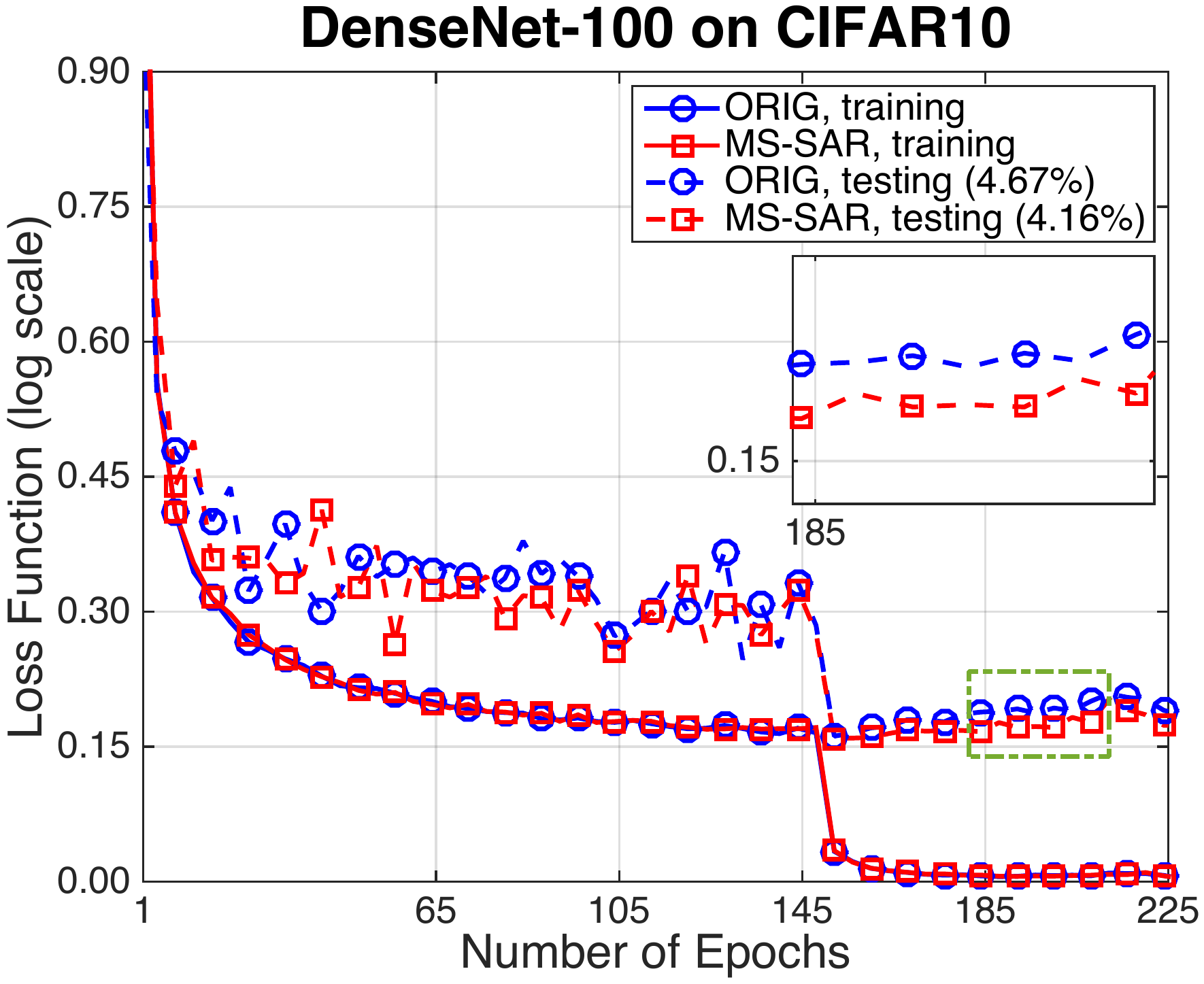}\\
    \includegraphics[width=4.0cm]{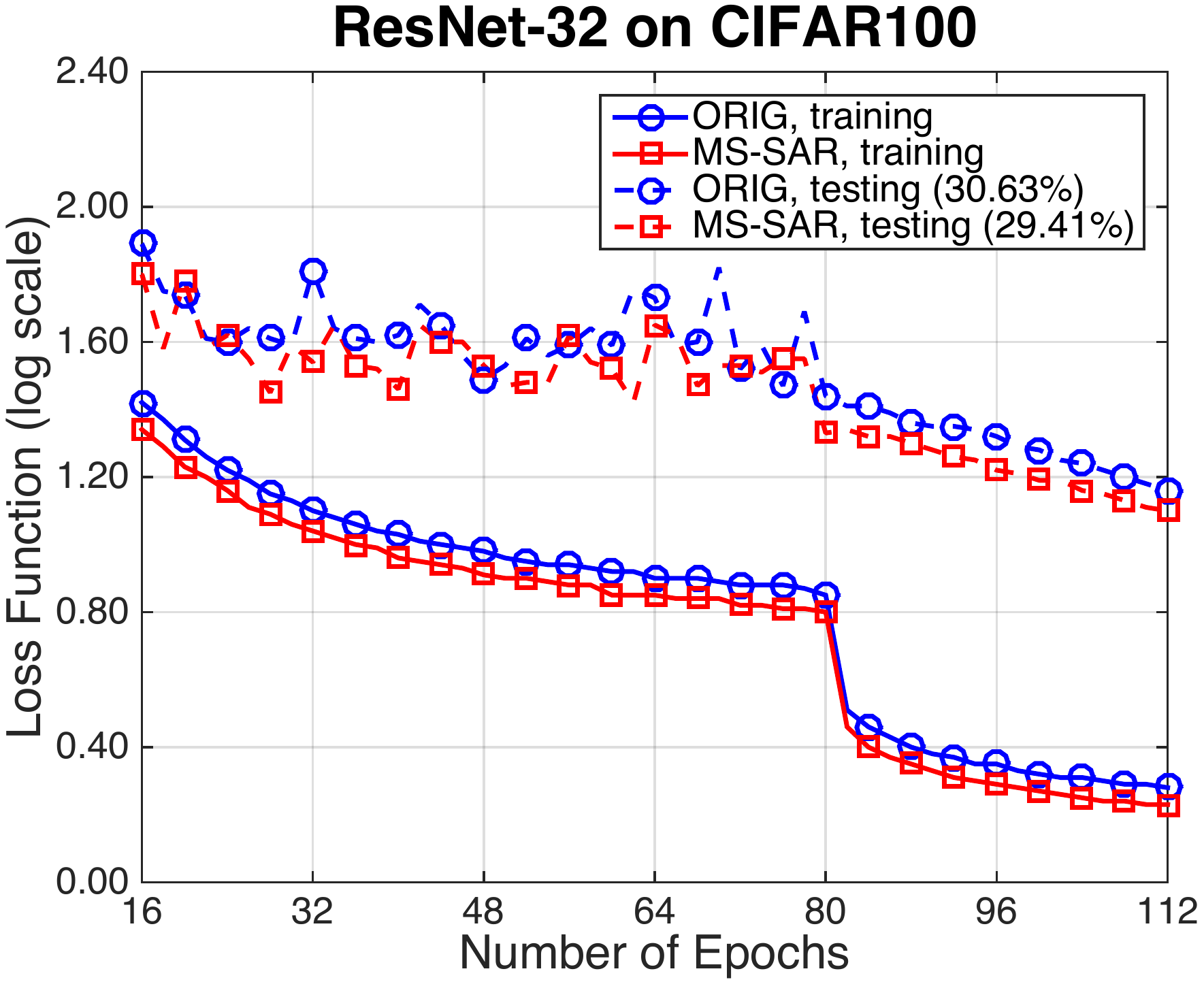}
    \includegraphics[width=4.0cm]{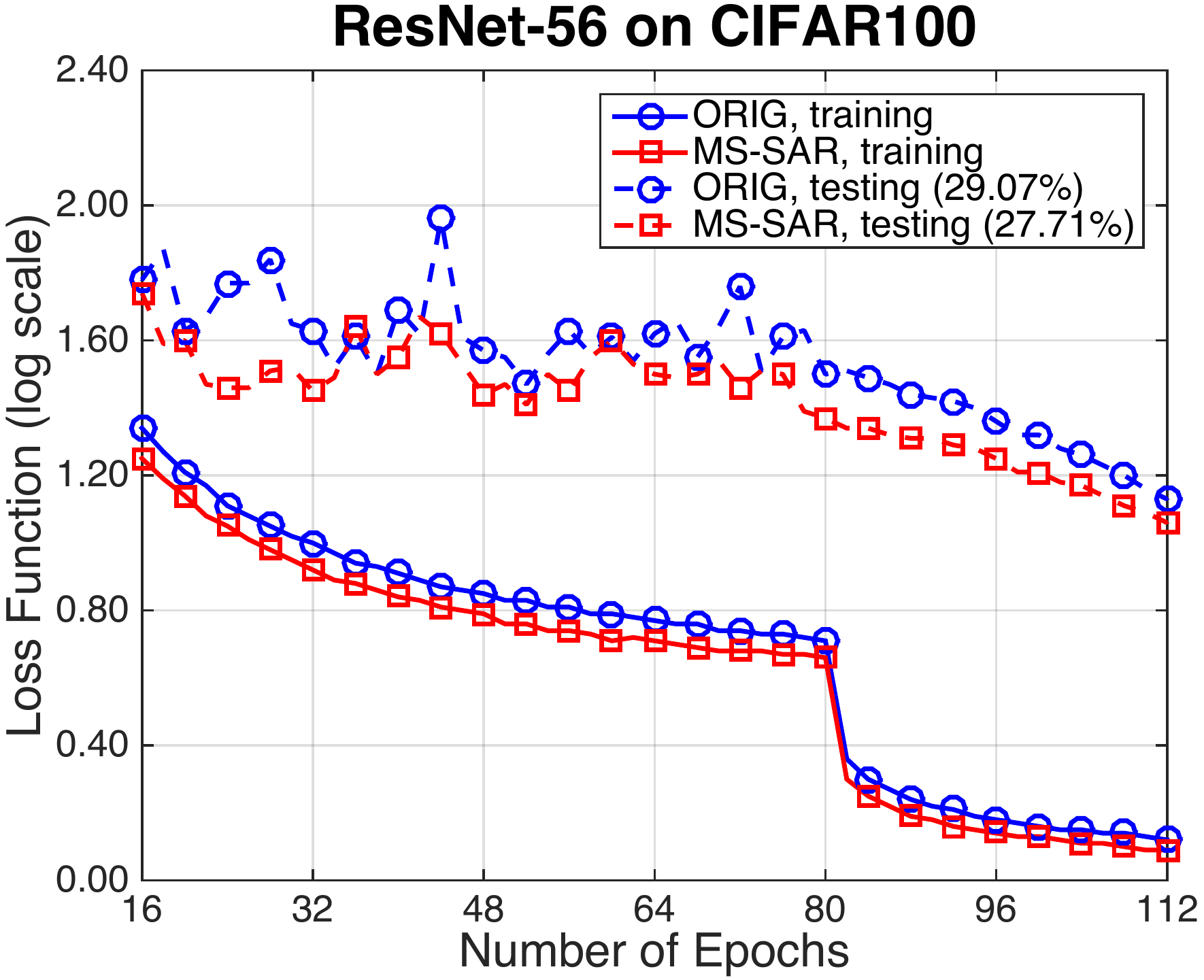}
    \includegraphics[width=4.0cm]{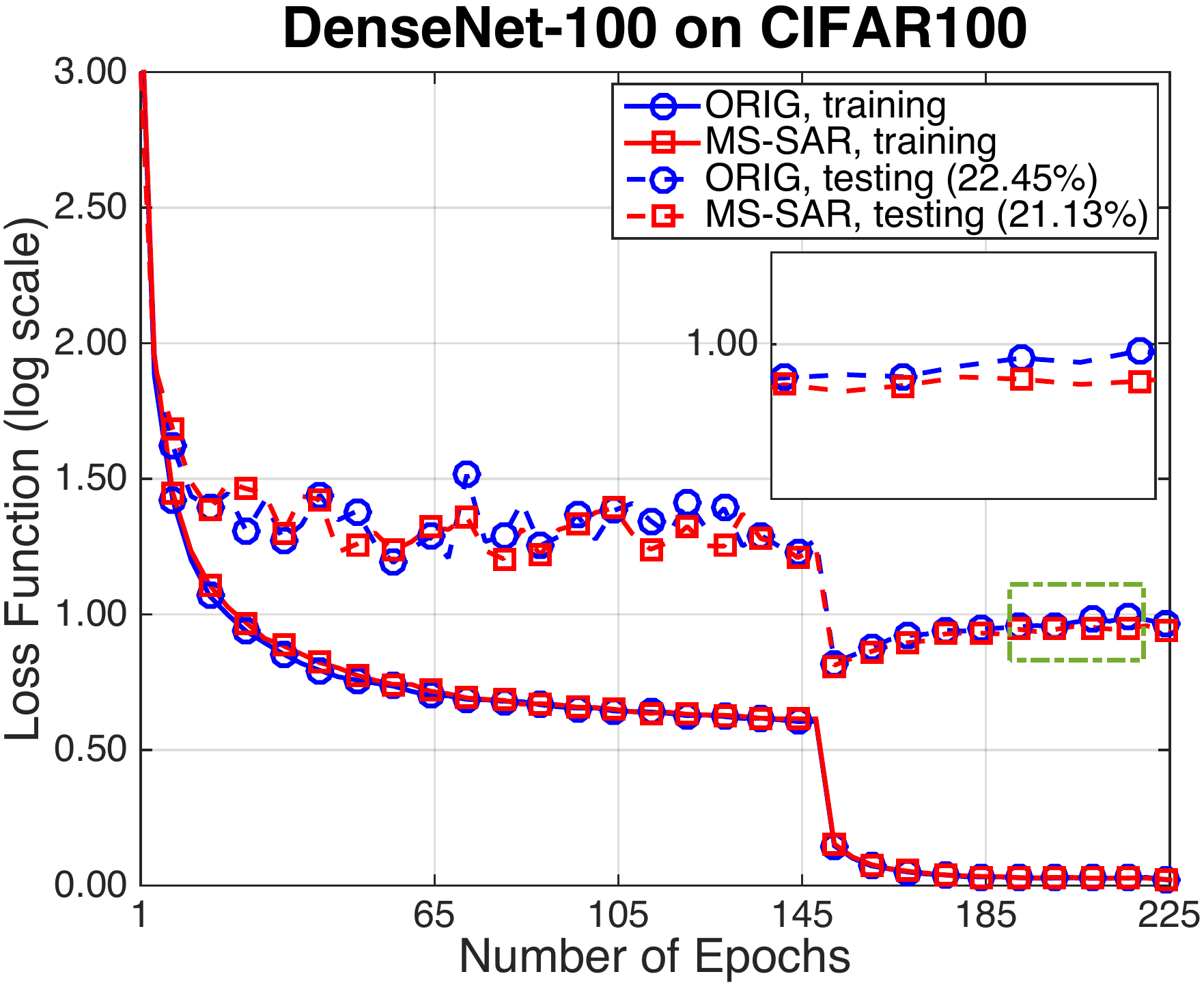}\\
\end{center}
\vspace{-0.2cm}
\caption{
    The curves of different networks (ResNet-32, ResNet-56 and DenseNet-100) with and without MS-SAR. All the curves on ResNet-32 and ResNet-56 are averaged over $3$ individual runs.
}
\label{Fig:CurvesCIFAR}
\end{figure}

In Figure~\ref{Fig:CurvesCIFAR}, we plot the training/testing curves of different networks on the CIFAR datasets. We find that MS-SAR effectively decreases the testing losses (and consequently, error rates) in all cases. On CIFAR10, due to the simplicity of the recognition task ($10$ classes), the training losses of both approaches, with and without MS-SAR, are very close to $0$, but MS-SAR produces lower testing losses, giving evidence for its ability to alleviate over-fitting.

\subsection{Ablation Study and Analysis}
\label{Experiments:Ablation}

\renewcommand{\colwidthA}{0.4cm}
\renewcommand{\colwidthB}{0.8cm}
\renewcommand{\colwidthC}{1.0cm}
\renewcommand{\colwidthD}{1.2cm}
\newcommand{\colwidthE}{1.3cm}
\begin{table*}[!btp]
\centering
\begin{tabular}{|C{0.4cm}|C{0.4cm}|C{0.4cm}||R{1.8cm}|R{2.0cm}|R{1.2cm}||R{0.8cm}|R{1.0cm}|R{1.2cm}|R{1.3cm}|}
\hline
\multicolumn{3}{|c||}{Scale}         &                                       \multicolumn{3}{c||}{ResNet-56} &                                          \multicolumn{4}{c|}{DenseNet-100} \\
\hline
$1$        & $2$        & $4$        &          C10 ($\pm$std) &         C100 ($\pm$std) &             FLOPs &              C10 &             C100 &             FLOPs &           Params \\
\hline\hline
           &            &            &          $ 6.97\pm0.05$ &          $29.41\pm0.14$ & $125.7\mathrm{M}$ &           $4.67$ &          $22.45$ & $252.5\mathrm{M}$ & $0.80\mathrm{M}$ \\
\hline
\checkmark &            &            &          $ 6.80\pm0.06$ &          $28.99\pm0.15$ & $125.7\mathrm{M}$ &           $4.45$ &          $21.83$ & $252.6\mathrm{M}$ & $0.99\mathrm{M}$ \\
\hline
           & \checkmark &            &          $ 6.55\pm0.05$ &          $28.31\pm0.17$ & $125.8\mathrm{M}$ &           $4.35$ &          $21.33$ & $253.0\mathrm{M}$ & $0.99\mathrm{M}$ \\
\hline
           &            & \checkmark &          $ 6.48\pm0.06$ &          $28.74\pm0.18$ & $126.3\mathrm{M}$ &           $4.39$ &          $21.79$ & $254.3\mathrm{M}$ & $0.99\mathrm{M}$ \\
\hline
\checkmark & \checkmark &            &          $ 6.38\pm0.07$ &          $28.28\pm0.19$ & $125.8\mathrm{M}$ &           $4.29$ &          $21.42$ & $252.8\mathrm{M}$ & $0.99\mathrm{M}$ \\
\hline
\checkmark &            & \checkmark &          $ 6.11\pm0.14$ &          $28.05\pm0.22$ & $126.0\mathrm{M}$ &           $4.32$ &          $21.27$ & $253.5\mathrm{M}$ & $0.99\mathrm{M}$ \\
\hline
           & \checkmark & \checkmark &          $ 6.35\pm0.09$ &          $28.87\pm0.27$ & $126.1\mathrm{M}$ &           $4.33$ &          $21.23$ & $253.7\mathrm{M}$ & $0.99\mathrm{M}$ \\
\hline
\checkmark & \checkmark & \checkmark & $\mathbf{ 6.04\pm0.11}$ & $\mathbf{27.71\pm0.21}$ & $126.0\mathrm{M}$ & $\mathbf{ 4.06}$ & $\mathbf{21.13}$ & $253.3\mathrm{M}$ & $0.99\mathrm{M}$ \\
\hline
\end{tabular}
\caption{
    Comparison of classification error rates ($\%$) on the CIFAR10 and CIFAR100 datasets with different scale combinations. Other specifications remain the same as in Figure~\ref{Tab:ResultsCIFAR}. All results of ResNet-56 are averaged over $3$ individual runs. See Section~\ref{Approach:Costs} for the reason that different scale configurations have the same number of parameters.
}
\label{Tab:ScaleCombinations}
\end{table*}

We first investigate the impacts of incorporating multi-scale visual information. To this end, we set $\mathcal{L}$ to be a non-empty subset of $\left\{1,2,4\right\}$ ($7$ possibilities), and summarize the results in Table~\ref{Tab:ScaleCombinations}. Compared with using a single scale, incorporating multi-scale information often leads to better classification performance (the only exception is that on DenseNet-100, ${\mathcal{L}}={\left\{2,4\right\}}$ works worse than ${\mathcal{L}}={\left\{2\right\}}$, which may be caused by random noise as DenseNet-100 experiments are performed only once). Combining all three scales is always produces the best recognition performance. Provided that the extra computational costs brought by multi-scale recalibration are almost ignorable, we will use ${\mathcal{L}}={\left\{1,2,4\right\}}$ in all the remaining experiments.

Next, we compare the two ways of defining coordinate sets ({\em sliding} vs. {\em regional}, see Section~\ref{Approach:Formulation}). In the experiments on CIFAR100, in both ResNets and DenseNets, the {\em regional} strategy outperforms the {\em sliding} strategy by $\sim0.2\%$. The {\em training} accuracy using the {\em sliding} strategy is also decreased, giving evidence that it is less capable of fitting training data. This reveals that, although spatial asymmetry is a nice property, its degree of freedom should be controlled, so that MS-SAR, containing a limited number of parameters, does not need to fit an over-complicated distribution. Considering that the {\em regional} strategy requires fewer computational costs (see Section~\ref{Approach:Costs}), we set it to be the default option.

Finally, we compare the {\em single-level} and {\em multi-level} recalibration methods on DenseNet-100. Detailed descriptions are in Section~\ref{Approach:Applications}. Note that this is independent of the comparison between {\em multi-scale} and {\em single-scale} methods -- they work on the spatial domain and the channel domain, and are complementary to each other. In the $100$-layer DenseNet, {\em multi-level} recalibration produces $4.06\%$ and $21.13\%$ error rates on CIFAR10 and CIFAR100, and these numbers are $4.45\%$ and $21.83\%$ for {\em single-level} recalibration, respectively. {\em Multi-level} recalibration reduces the relative errors by $7.77\%$ and $5.12\%$, at the price of $23.75\%$ extra parameters and $0.3\%$ additional FLOPs.

\subsection{The ILSVRC2012 Dataset}
\label{Experiments:ILSVRC2012}

\begin{table*}[!btp]
\centering
\begin{tabular}{|c||c||c|c|c|}
\hline
Layer        & Size           & ResNet-18                 & ResNet-34                 & ResNeXt-50                \\
\hline\hline
{\em conv-1} & $224\times224$ & \multicolumn{3}{c|}{$7\times7$, $64$ (kernel size, channel \#)}                   \\
\hline
{\em pool-1} & $112\times112$ & \multicolumn{3}{c|}{max-pooling, $3\times3$, a stride of $2$}                     \\
\hline
{\em conv-2} & $ 56\times 56$ & $\begin{bmatrix} 3\times3,   64       \\ 3\times3,   64 \end{bmatrix}\times 2$
                              & $\begin{bmatrix} 3\times3,   64       \\ 3\times3,   64 \end{bmatrix}\times 3$
                              & $\begin{bmatrix} 1\times1,   64, G=32 \\ 3\times3,   64, G=32 \\ 1\times1,  256, G=32 \end{bmatrix}\times 3$ \\
\hline
{\em conv-3} & $ 28\times 28$ & $\begin{bmatrix} 3\times3,  128       \\ 3\times3,  128 \end{bmatrix}\times 2$
                              & $\begin{bmatrix} 3\times3,  128       \\ 3\times3,  128 \end{bmatrix}\times 4$
                              & $\begin{bmatrix} 1\times1,  128, G=32 \\ 3\times3,  128, G=32 \\ 1\times1,  512, G=32 \end{bmatrix}\times 4$ \\
\hline
{\em conv-4} & $ 14\times 14$ & $\begin{bmatrix} 3\times3,  256       \\ 3\times3,  256 \end{bmatrix}\times 2$
                              & $\begin{bmatrix} 3\times3,  256       \\ 3\times3,  256 \end{bmatrix}\times 6$
                              & $\begin{bmatrix} 1\times1,  256, G=32 \\ 3\times3,  256, G=32 \\ 1\times1, 1024, G=32 \end{bmatrix}\times 6$ \\
\hline
{\em conv-5} & $  7\times  7$ & $\begin{bmatrix} 3\times3,  512       \\ 3\times3,  512 \end{bmatrix}\times 2$
                              & $\begin{bmatrix} 3\times3,  512       \\ 3\times3,  512 \end{bmatrix}\times 3$
                              & $\begin{bmatrix} 1\times1,  512, G=32 \\ 3\times3,  512, G=32 \\ 1\times1, 2048, G=32 \end{bmatrix}\times 3$ \\
\hline
{\em fc}     & $  1\times  1$ & \multicolumn{3}{c|}{global average pooling, $1\times1$, $1000$, fully-connected}  \\
\hline\hline
FLOPs         &              & $1.81\mathrm{G}$           & $3.66\mathrm{G}$          & $3.86\mathrm{G}$          \\
\hline
\end{tabular}
\caption{
    Network Architectures for the ILSVRC2012 dataset. Batch normalization~\cite{ioffe2015batch} and ReLU activation~\cite{nair2010rectified} are used as in the original papers~\cite{he2016deep}\cite{xie2017aggregated}. ${G}={32}$ means to use $32$ subgroups in convolution, as suggested in~\cite{xie2017aggregated}.
}
\label{Tab:ArchitecturesILSVRC2012}
\end{table*}

The ILSVRC2012 dataset~\cite{russakovsky2015imagenet} is a subset of the ImageNet database~\cite{deng2009imagenet}, created for a large-scale visual recognition competition. It contains $1\rm{,}000$ categories located at different levels of the WordNet hierarchy. The training set has $\sim1.3\mathrm{M}$ images, which are roughly uniformly distributed over all classes. The testing set ({\em a.k.a.}, the validation set used in the competition) has $50\mathrm{K}$ images, or exactly $50$ images for each class.

The baseline network architectures include two ResNets~\cite{he2016deep} with $18$ and $34$ layers, and a ResNeXt~\cite{xie2017aggregated} with $50$ layers. Detailed network architectures are shown in Table~\ref{Tab:ArchitecturesILSVRC2012}. We also compare with the Squeeze-and-Excitation (SE) module~\cite{hu2017squeeze}, which is a special case of our approach (${\mathcal{L}}={\left\{1\right\}}$, which is single-scale and spatially-symmetric). As illustrated in Figure~\ref{Fig:Examples}, both SE and MS-SAR modules are appended after the second convolutional layer of each residual block.

All these networks are trained from scratch. We follow~\cite{hu2017squeeze} in configuring the following parameters. Standard Stochastic Gradient Descent (SGD) with a weight decay of $0.0001$ and a Nesterov momentum of $0.9$ is used. There are a total of $100$ epochs in the training process, and the mini-batch size is $1024$. The learning rate starts with $0.6$, and is divided by $10$ after $30$, $60$ and $90$ epochs. Again, adding MS-SAR does not require any of these settings to be modified. In the training process, we apply a series of data-augmentation techniques, including rescaling and cropping the image, randomly mirroring and rotating (slightly) the image, changing its aspect ratio and performing pixel jittering, which is same with SENet\cite{hu2017squeeze}. In the testing process, we use the standard single-center-crop on each image.

\renewcommand{\colwidthA}{0.4cm}
\renewcommand{\colwidthB}{1.0cm}
\renewcommand{\colwidthC}{1.0cm}
\renewcommand{\colwidthD}{1.2cm}
\renewcommand{\colwidthE}{1.3cm}
\begin{table*}[!btp]
\centering
\begin{tabular}{|l||C{\colwidthA}|C{\colwidthA}|C{\colwidthA}||R{\colwidthB}|R{\colwidthC}|R{\colwidthD}|R{\colwidthE}|}
\hline
\multirow{2}{*}{Approach} & \multicolumn{3}{|c||}{Scale}         & \multirow{2}{*}{Top-$1$} & \multirow{2}{*}{Top-$5$} & \multirow{2}{*}{FLOPs} & \multirow{2}{*}{Params} \\
\cline{2-4}
                          & $1$        & $2$        & $4$        &                          &                          &                        &                         \\
\hline\hline
ResNet-18                 &            &            &            &                  $30.50$ &                  $11.07$ &        $1.81\mathrm{G}$ &        $10.9\mathrm{M}$ \\
\hline
ResNet-18$+$SE            & \checkmark &            &            &                  $29.78$ &                  $10.27$ &        $1.81\mathrm{G}$ &        $13.8\mathrm{M}$ \\
\hline
ResNet-18$+$MS-SAR        & \checkmark & \checkmark & \checkmark &         $\mathbf{29.43}$ &         $\mathbf{10.19}$ &        $1.81\mathrm{G}$ &        $13.8\mathrm{M}$ \\
\hline\hline
ResNet-34                 &            &            &            &                  $27.02$ &                  $ 8.77$ &        $3.66\mathrm{G}$ &        $21.7\mathrm{M}$ \\
\hline
ResNet-34$+$SE            & \checkmark &            &            &                  $26.67$ &                  $ 8.43$ &        $3.66\mathrm{G}$ &        $27.3\mathrm{M}$ \\
\hline
ResNet-34$+$MS-SAR        & \checkmark & \checkmark & \checkmark &         $\mathbf{26.15}$ &         $\mathbf{8.35}$ &        $3.67\mathrm{G}$ &        $27.4\mathrm{M}$ \\
\hline\hline
ResNeXt-50                &            &            &            &                  $22.20$ &                  $ 6.12$ &        $3.86\mathrm{G}$ &        $25.0\mathrm{M}$ \\
\hline
ResNeXt-50$+$SE           & \checkmark &            &            &                  $21.95$ &                  $ 5.93$ &        $3.87\mathrm{G}$ &        $27.5\mathrm{M}$ \\
\hline
ResNeXt-50$+$MS-SAR       & \checkmark & \checkmark & \checkmark &         $\mathbf{21.64}$ &         $\mathbf{ 5.78}$ &        $3.89\mathrm{G}$ &        $27.6\mathrm{M}$ \\
\hline
\end{tabular}
\caption{
    Comparison of top-$1$ and top-$5$ classification error rates ($\%$) produced by different recalibration approaches (none, SE and MS-SAR) on the ILSVRC2012 dataset. All these numbers are based on our own implementation. See Section~\ref{Approach:Costs} for the reason that different scale configurations have the same number of parameters.
}
\label{Tab:ResultsILSVRC2012}
\end{table*}

\begin{figure}[t]
\begin{center}
    \includegraphics[width=5.6cm]{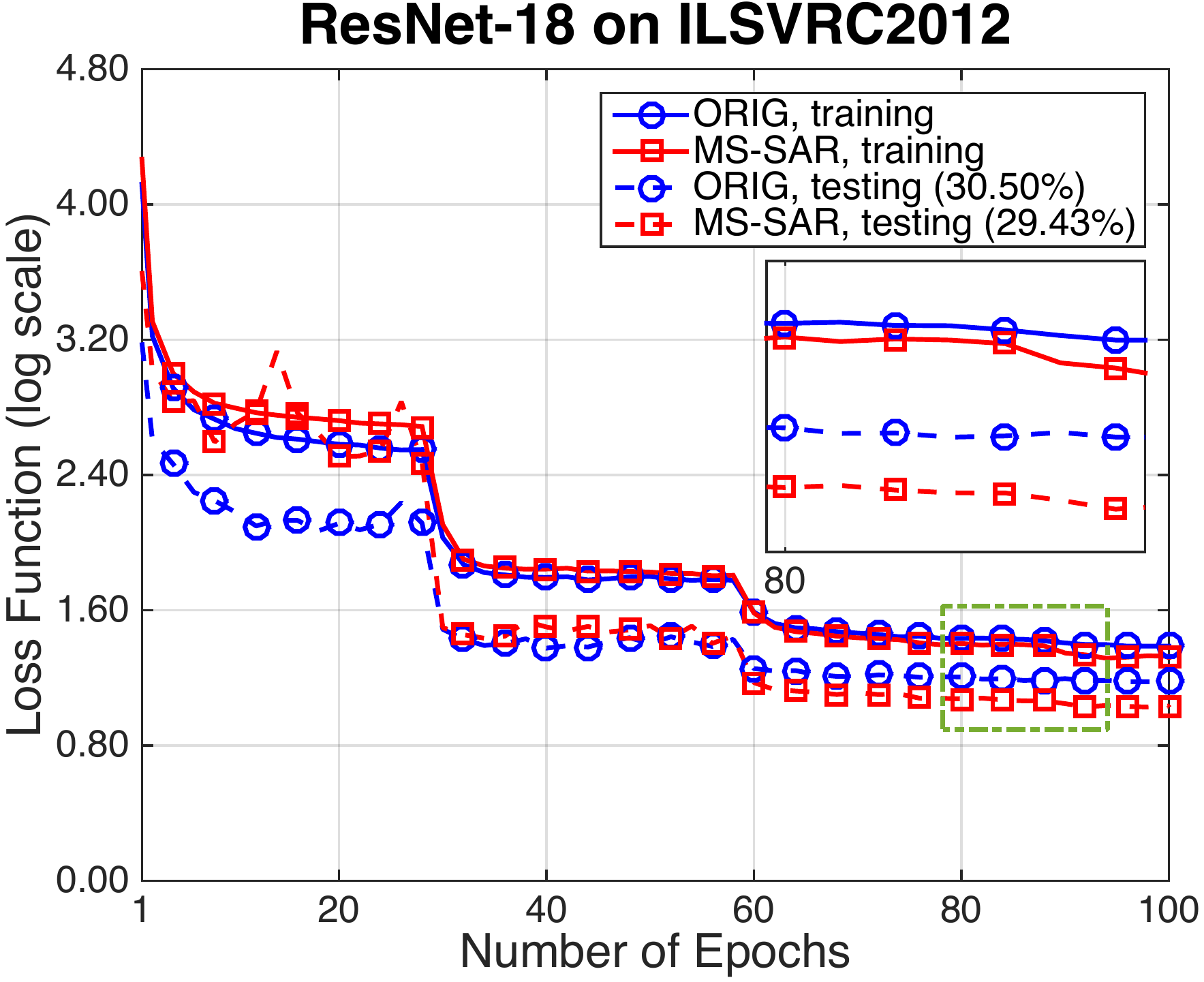}
    \includegraphics[width=5.6cm]{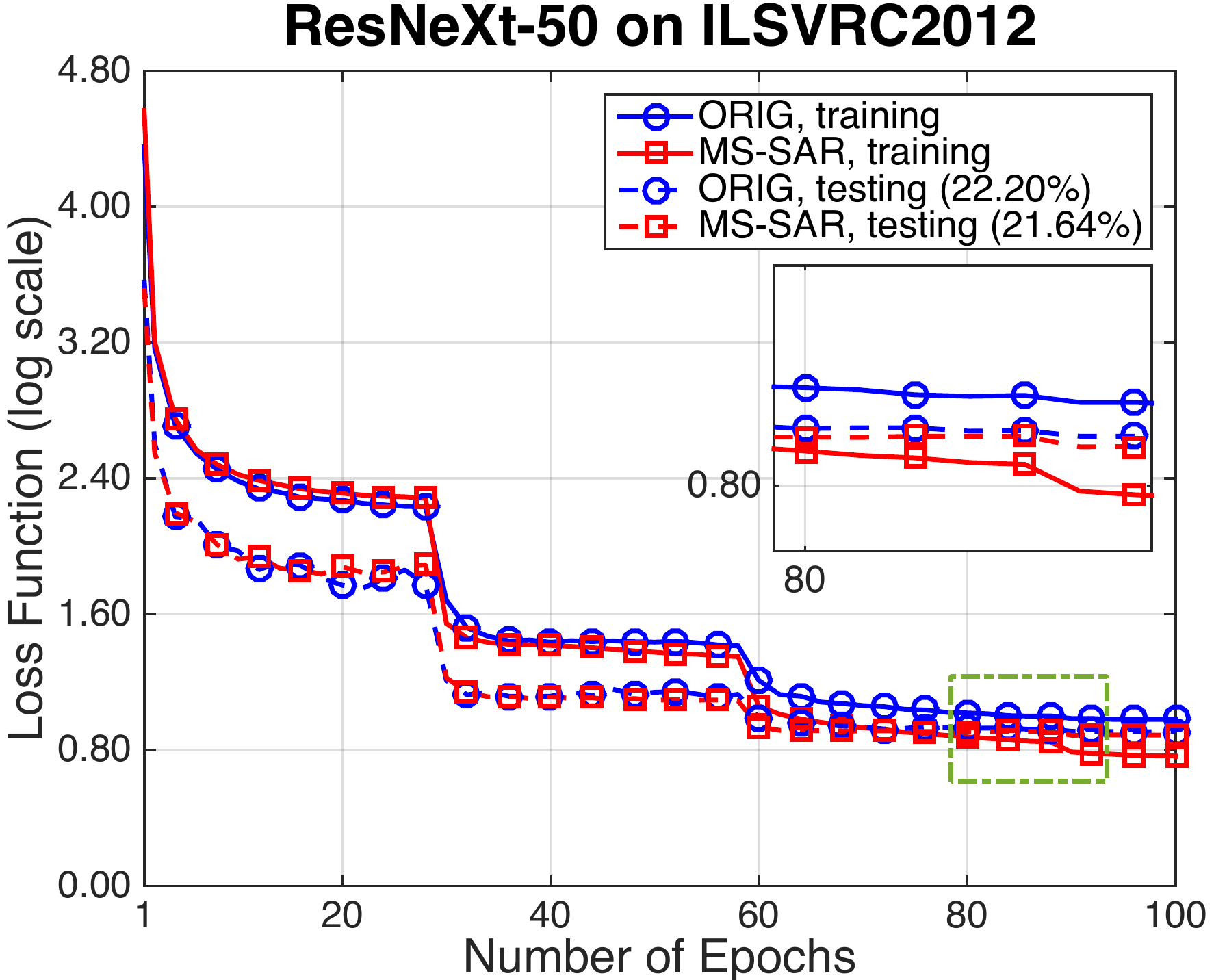}\\
\end{center}
\caption{
    The curves of different networks with and without MS-SAR on the ILSVRC2012 dataset. We zoom-in on a small part of each curve for better visualization.
}
\label{Fig:CurvesILSVRC2012}
\end{figure}

Results are summarized in Table~\ref{Tab:ResultsILSVRC2012}. In all cases, MS-SAR works better than the baseline (no recalibration) and SE (single-scale spatially-symmetric recalibration). For example, based on ResNeXt-50, MS-SAR reduces the top-5 error of the baseline by an absolute value of $0.34\%$ or a relative value of $5.56\%$, using $\sim1\%$ extra FLOPs and $~10\%$ extra parameters. On top of SE, the error rate drops are $0.15\%$ (absolute) and $2.53\%$ (relative) and the extra FLOPs and parameters are merely $\sim0.5\%$ and $\sim0.4\%$, respectively. The training/testing curves in Figure~\ref{Fig:CurvesILSVRC2012} show similar phenomena as in CIFAR experiments.

\begin{figure}[t]
\begin{center}
    \includegraphics[width=3.8cm]{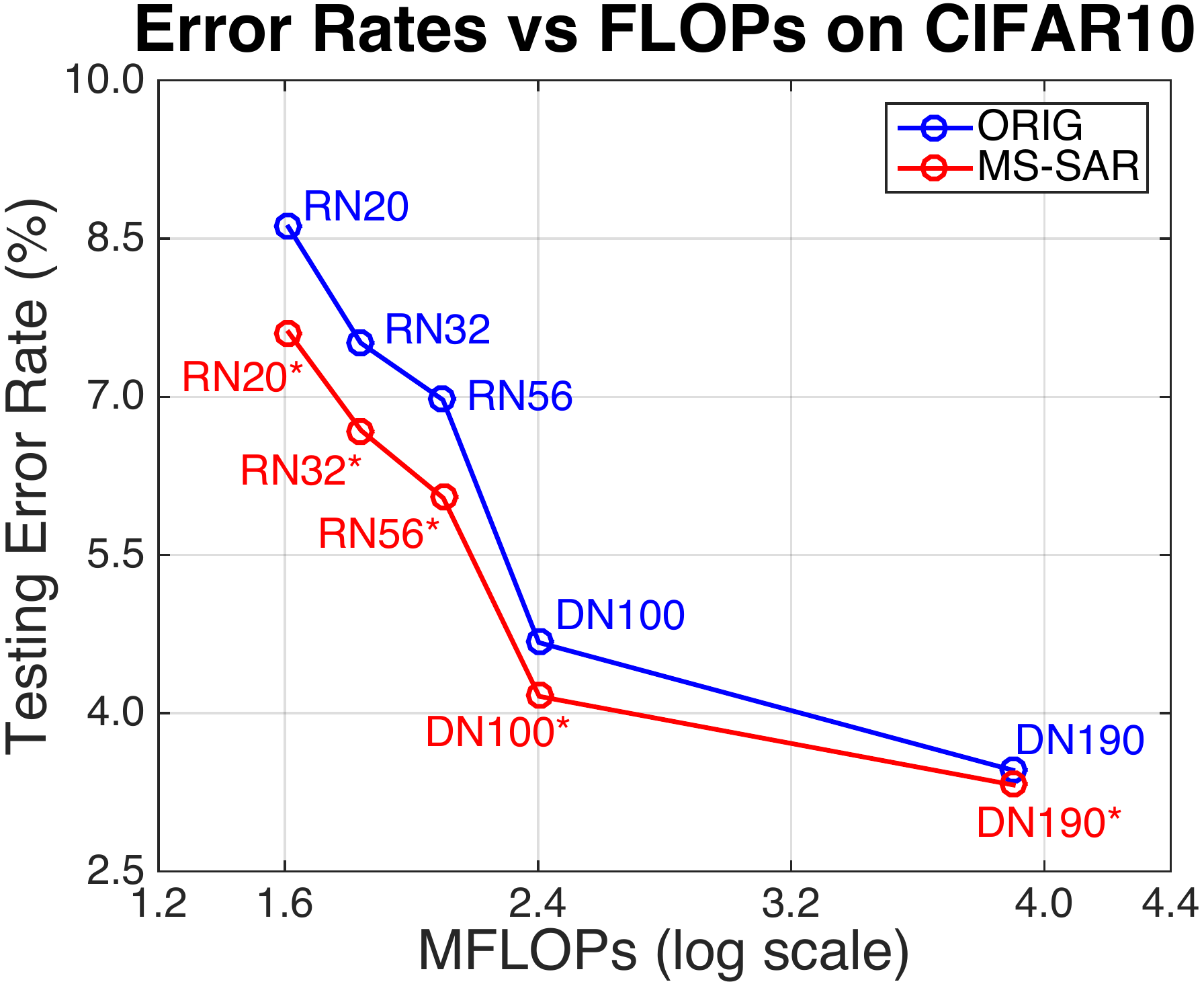}
    \includegraphics[width=3.8cm]{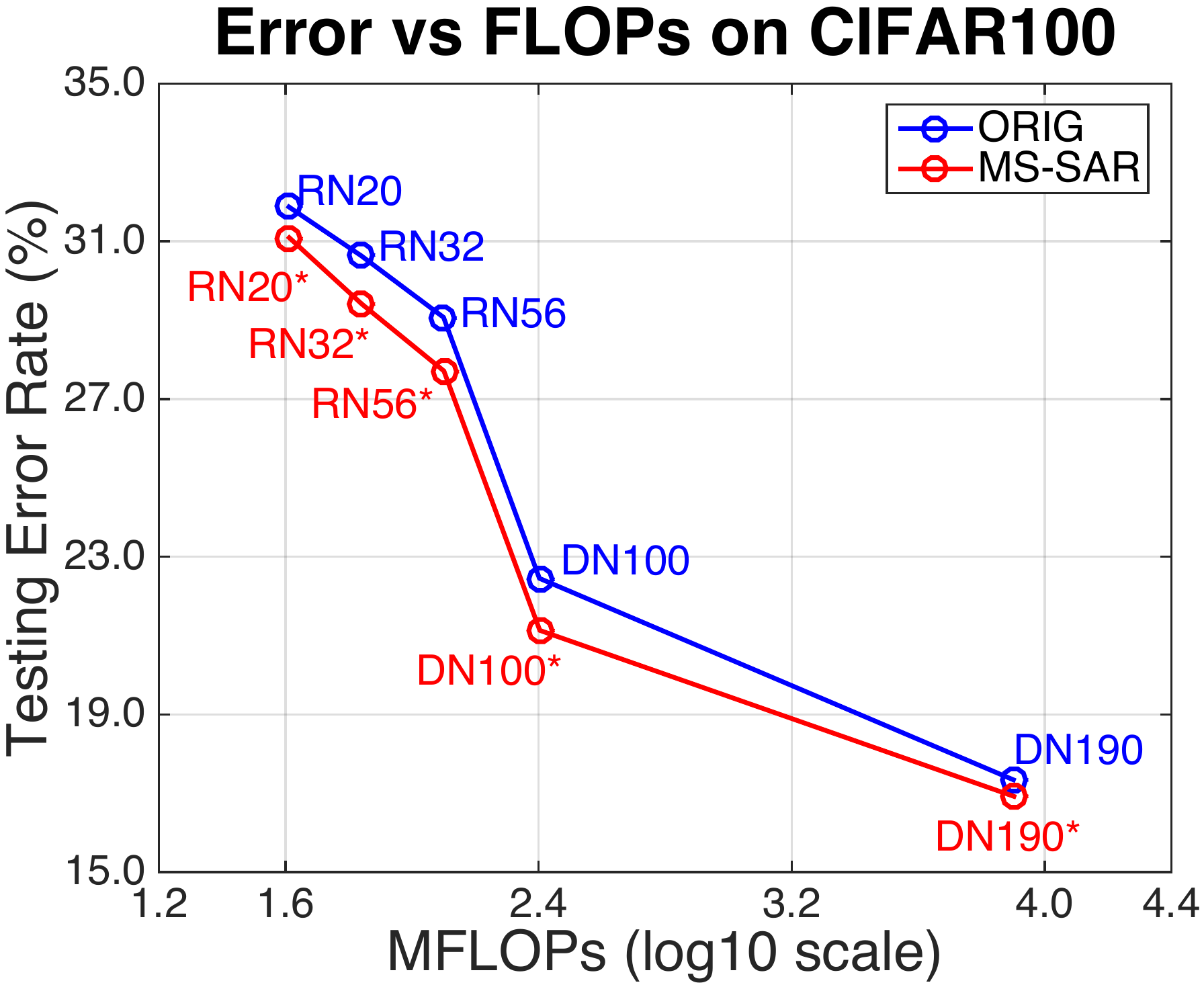}
    \includegraphics[width=3.8cm]{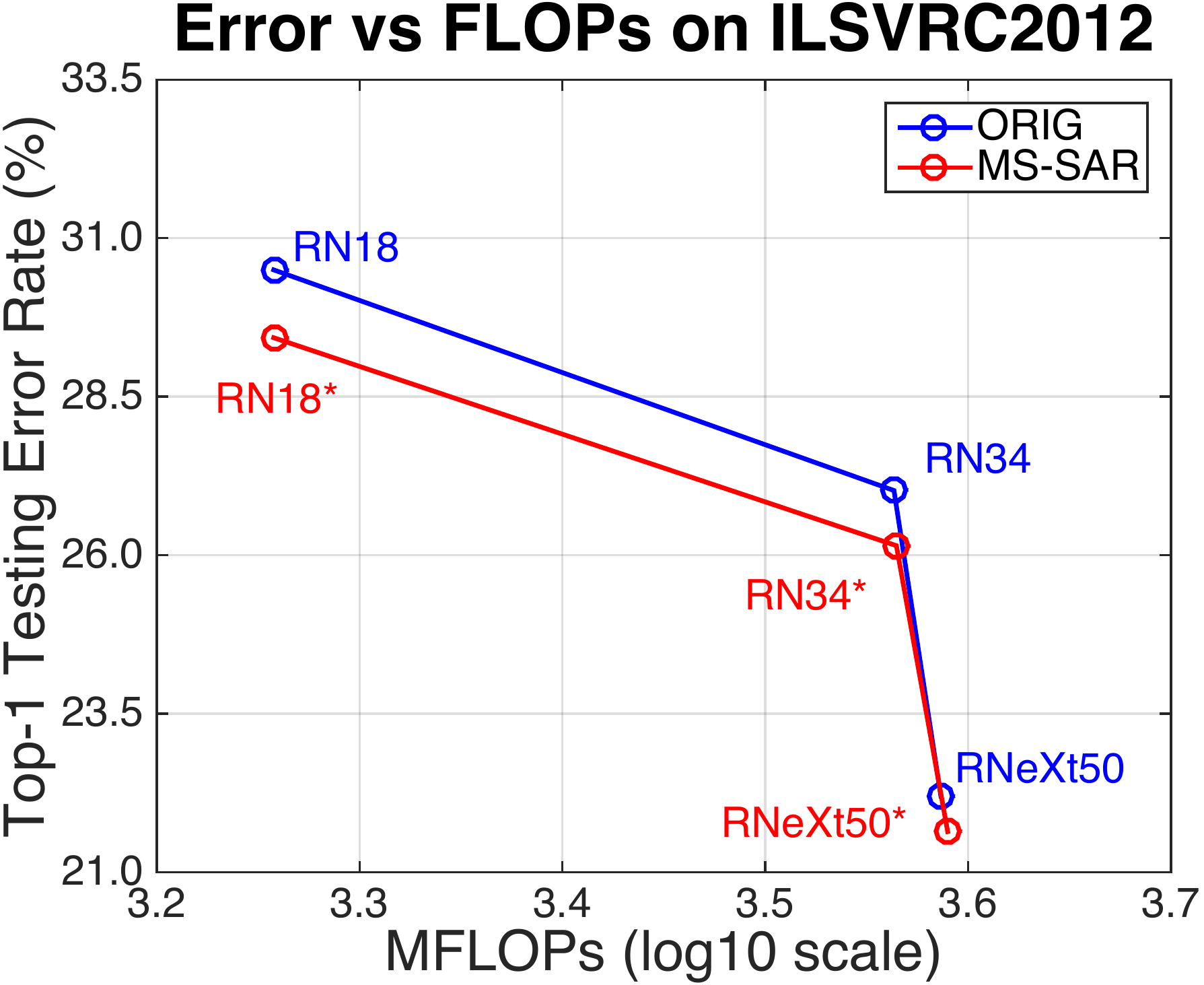}\\
\end{center}
\caption{
    The relationship between classification accuracy and computation (in FLOPs) on three datasets. RN, DN and RNeXt denote ResNet, DenseNet and ResNeXt, respectively. An asterisk sign (*) indicates that MS-SAR is added.
}
\label{Fig:Computations}
\end{figure}

Lastly, we investigate the relationship between classification accuracy and computation on these three datasets. In Figure~\ref{Fig:Computations}, we plot the testing error as the function of FLOPs, which reveals the trend that MS-SAR can achieve higher recognition accuracy under the same computational complexity.

\section{Conclusions}
\label{Conclusions}

In this paper, we present the multi-scale spatially-asymmetric recalibration (MS-SAR) module for image classification. This is aimed at assigning the convolutional layer with the ability to incorporate spatial contexts to ``recalibrate'' each neural response, {\em i.e.}, summarizing regional information into an importance factor and multiplying it to the original response. We implement each recalibration function as the combination of a pooling operation in the spatial domain and a linear model in the channel domain. We also utilize multi-scale information to estimate the importance of each neuron more accurately. Experiments on the CIFAR and ILSVRC2012 datasets demonstrate the superior performance of MS-SAR over several baseline network architectures.

Our work delivers two messages. First, it is not the best choice to rely on a gradually increasing receptive field (via local convolution, pooling or down-sampling) to capture spatial information -- MS-SAR is a light-weighted yet specifically designed module which deals with this issue more efficiently. Second, there exists a tradeoff between diversity and simplicity -- this is why {\em regional} pooling works better than {\em sliding} pooling. In its current form, MS-SAR is able to add a weight factor to each neural response (unary or linear terms), but unable to explicitly model the co-occurrence of multiple features (binary or higher-order terms). We leave this topic for future research.

\section*{Acknowledgements}
\label{Acknowledgements}
We thank Huiyu Wang for instructive discussions.

\bibliographystyle{splncs}
\bibliography{egbib}

\begin{thebibliography}{10}

\bibitem{krizhevsky2012imagenet}
Krizhevsky, A., Sutskever, I., Hinton, G.:
\newblock Imagenet classification with deep convolutional neural networks.
\newblock In: Advances in Neural Information Processing Systems. (2012)

\bibitem{girshick2015fast}
Girshick, R.:
\newblock Fast r-cnn.
\newblock In: Computer Vision and Pattern Recognition. (2015)

\bibitem{long2015fully}
Long, J., Shelhamer, E., Darrell, T.:
\newblock Fully convolutional networks for semantic segmentation.
\newblock In: Computer Vision and Pattern Recognition. (2015)

\bibitem{xie2015holistically}
Xie, S., Tu, Z.:
\newblock Holistically-nested edge detection.
\newblock In: International Conference on Computer Vision. (2015)

\bibitem{nair2010rectified}
Nair, V., Hinton, G.E.:
\newblock Rectified linear units improve restricted boltzmann machines.
\newblock In: International Conference on Machine Learning. (2010)

\bibitem{hu2017squeeze}
Hu, J., Shen, L., Sun, G.:
\newblock Squeeze-and-excitation networks.
\newblock arXiv preprint arXiv:1709.01507 (2017)

\bibitem{he2016deep}
He, K., Zhang, X., Ren, S., Sun, J.:
\newblock Deep residual learning for image recognition.
\newblock In: Computer Vision and Pattern Recognition. (2016)

\bibitem{huang2017densely}
Huang, G., Liu, Z., Weinberger, K.Q., van~der Maaten, L.:
\newblock Densely connected convolutional networks.
\newblock In: Computer Vision and Pattern Recognition. (2017)

\bibitem{krizhevsky2009learning}
Krizhevsky, A., Hinton, G.:
\newblock Learning multiple layers of features from tiny images.
\newblock (2009)

\bibitem{russakovsky2015imagenet}
Russakovsky, O., Deng, J., Su, H., Krause, J., Satheesh, S., Ma, S., Huang, Z.,
  Karpathy, A., Khosla, A., Bernstein, M.,  et~al.:
\newblock Imagenet large scale visual recognition challenge.
\newblock International Journal of Computer Vision \textbf{115}(3) (2015)
  211--252

\bibitem{lecun1998gradient}
LeCun, Y., Bottou, L., Bengio, Y., Haffner, P.:
\newblock Gradient-based learning applied to document recognition.
\newblock Proceedings of the IEEE \textbf{86}(11) (1998)  2278--2324

\bibitem{deng2009imagenet}
Deng, J., Dong, W., Socher, R., Li, L., Li, K., Fei-Fei, L.:
\newblock Imagenet: A large-scale hierarchical image database.
\newblock In: Computer Vision and Pattern Recognition. (2009)

\bibitem{lin2014microsoft}
Lin, T.Y., Maire, M., Belongie, S., Hays, J., Perona, P., Ramanan, D., Dollar,
  P., Zitnick, C.:
\newblock Microsoft coco: Common objects in context.
\newblock In: European conference on computer vision. (2014)

\bibitem{srivastava2014dropout}
Srivastava, N., Hinton, G.E., Krizhevsky, A., Sutskever, I., Salakhutdinov, R.:
\newblock Dropout: A simple way to prevent neural networks from overfitting.
\newblock Journal of Machine Learning Research \textbf{15}(1) (2014)
  1929--1958

\bibitem{wang2017sort}
Wang, Y., Xie, L., Liu, C., Qiao, S., Zhang, Y., Zhang, W., Tian, Q., Yuille,
  A.:
\newblock Sort: Second-order response transform for visual recognition.
\newblock In: International Conference on Computer Vision. (2017)

\bibitem{szegedy2015going}
Szegedy, C., Liu, W., Jia, Y., Sermanet, P., Reed, S., Anguelov, D., Erhan, D.,
  Vanhoucke, V., Rabinovich, A.,  et~al.:
\newblock Going deeper with convolutions.
\newblock In: Computer Vision and Pattern Recognition. (2015)

\bibitem{simonyan2015very}
Simonyan, K., Zisserman, A.:
\newblock Very deep convolutional networks for large-scale image recognition.
\newblock In: International Conference on Learning Representations. (2015)

\bibitem{chen2017dual}
Chen, Y., Li, J., Xiao, H., Jin, X., Yan, S., Feng, J.:
\newblock Dual path networks.
\newblock In: Advances in Neural Information Processing Systems. (2017)

\bibitem{ioffe2015batch}
Ioffe, S., Szegedy, C.:
\newblock Batch normalization: Accelerating deep network training by reducing
  internal covariate shift.
\newblock In: International Conference on Machine Learning. (2015)

\bibitem{srivastava2015highway}
Srivastava, R.K., Greff, K., Schmidhuber, J.:
\newblock Highway networks.
\newblock arXiv preprint arXiv:1505.00387 (2015)

\bibitem{xie2017genetic}
Xie, L., Yuille, A.:
\newblock Genetic cnn.
\newblock In: International Conference on Computer Vision. (2017)

\bibitem{zoph2017neural}
Zoph, B., Le, Q.V.:
\newblock Neural architecture search with reinforcement learning.
\newblock In: International Conference on Learning Representations. (2017)

\bibitem{donahue2014decaf}
Donahue, J., Jia, Y., Vinyals, O., Hoffman, J., Zhang, N., Tzeng, E., Darrell,
  T.:
\newblock Decaf: A deep convolutional activation feature for generic visual
  recognition.
\newblock In: International Conference on Machine Learning. (2014)

\bibitem{razavian2014cnn}
Razavian, A.S., Azizpour, H., Sullivan, J., Carlsson, S.:
\newblock Cnn features off-the-shelf: an astounding baseline for recognition.
\newblock In: Computer Vision and Pattern Recognition. (2014)

\bibitem{zhang2014part}
Zhang, N., Donahue, J., Girshick, R., Darrell, T.:
\newblock Part-based r-cnns for fine-grained category detection.
\newblock In: European Conference on Computer Vision. (2014)

\bibitem{xie2016interactive}
Xie, L., Zheng, L., Wang, J., Yuille, A., Tian, Q.:
\newblock Interactive: Inter-layer activeness propagation.
\newblock In: Computer Vision and Pattern Recognition. (2016)

\bibitem{girshick2014rich}
Girshick, R., Donahue, J., Darrell, T., Malik, J.:
\newblock Rich feature hierarchies for accurate object detection and semantic
  segmentation.
\newblock In: Computer Vision and Pattern Recognition. (2014)

\bibitem{ren2015faster}
Ren, S., He, K., Girshick, R., Sun, J.:
\newblock Faster r-cnn: Towards real-time object detection with region proposal
  networks.
\newblock In: Advances in Neural Information Processing Systems. (2015)

\bibitem{chen2016deeplab}
Chen, L.C., Papandreou, G., Kokkinos, I., Murphy, K., Yuille, A.L.:
\newblock Deeplab: Semantic image segmentation with deep convolutional nets,
  atrous convolution, and fully connected crfs.
\newblock In: International Conference on Learning Representations. (2016)

\bibitem{toshev2014deeppose}
Toshev, A., Szegedy, C.:
\newblock Deeppose: Human pose estimation via deep neural networks.
\newblock In: Computer Vision and Pattern Recognition. (2014)

\bibitem{newell2016stacked}
Newell, A., Yang, K., Deng, J.:
\newblock Stacked hourglass networks for human pose estimation.
\newblock In: European Conference on Computer Vision. (2016)

\bibitem{he2014spatial}
He, K., Zhang, X., Ren, S., Sun, J.:
\newblock Spatial pyramid pooling in deep convolutional networks for visual
  recognition.
\newblock In: European Conference on Computer Vision. (2014)

\bibitem{zeiler2014visualizing}
Zeiler, M.D., Fergus, R.:
\newblock Visualizing and understanding convolutional networks.
\newblock In: European Conference on Computer Vision. (2014)

\bibitem{simonyan2013deep}
Simonyan, K., Vedaldi, A., Zisserman, A.:
\newblock Deep inside convolutional networks: Visualising image classification
  models and saliency maps.
\newblock arXiv preprint arXiv:1312.6034 (2013)

\bibitem{noh2015learning}
Noh, H., Hong, S., Han, B.:
\newblock Learning deconvolution network for semantic segmentation.
\newblock In: International Conference on Computer Vision. (2015)

\bibitem{chen2016attention}
Chen, L.C., Yang, Y., Wang, J., Xu, W., Yuille, A.L.:
\newblock Attention to scale: Scale-aware semantic image segmentation.
\newblock In: Computer Vision and Pattern Recognition. (2016)

\bibitem{xie2016geometric}
Xie, L., Tian, Q., Flynn, J., Wang, J., Yuille, A.:
\newblock Geometric neural phrase pooling: Modeling the spatial co-occurrence
  of neurons.
\newblock In: European Conference on Computer Vision. (2016)

\bibitem{lee2015deeply}
Lee, C., Xie, S., Gallagher, P., Zhang, Z., Tu, Z.:
\newblock Deeply-supervised nets.
\newblock In: Artificial Intelligence and Statistics. (2015)

\bibitem{huang2016deep}
Huang, G., Sun, Y., Liu, Z., Sedra, D., Weinberger, K.Q.:
\newblock Deep networks with stochastic depth.
\newblock In: European Conference on Computer Vision. (2016)

\bibitem{he2016identity}
He, K., Zhang, X., Ren, S., Sun, J.:
\newblock Identity mappings in deep residual networks.
\newblock In: European Conference on Computer Vision. (2016)

\bibitem{zagoruyko2016wide}
Zagoruyko, S., Komodakis, N.:
\newblock Wide residual networks.
\newblock arXiv preprint arXiv:1605.07146 (2016)

\bibitem{han2017deep}
Han, D., Kim, J., Kim, J.:
\newblock Deep pyramidal residual networks.
\newblock In: Computer Vision and Pattern Recognition. (2017)

\bibitem{huang2017snapshot}
Huang, G., Li, Y., Pleiss, G., Liu, Z., Hopcroft, J.E., Weinberger, K.Q.:
\newblock Snapshot ensembles: Train 1, get m for free.
\newblock In: International Conference on Learning Representations. (2017)

\bibitem{zhang2017interleaved}
Zhang, T., Qi, G.J., Xiao, B., Wang, J.:
\newblock Interleaved group convolutions.
\newblock In: Computer Vision and Pattern Recognition. (2017)

\bibitem{gastaldi2017shake}
Gastaldi, X.:
\newblock Shake-shake regularization.
\newblock arXiv preprint arXiv:1705.07485 (2017)

\bibitem{zhang2017mixup}
Zhang, H., Cisse, M., Dauphin, Y.N., Lopez-Paz, D.:
\newblock mixup: Beyond empirical risk minimization.
\newblock arXiv preprint arXiv:1710.09412 (2017)

\bibitem{xie2017aggregated}
Xie, S., Girshick, R., Dollar, P., Tu, Z., He, K.:
\newblock Aggregated residual transformations for deep neural networks.
\newblock In: Computer Vision and Pattern Recognition. (2017)

\end{thebibliography}
\end{document}